\documentclass[journal]{IEEEtran}

\usepackage{times}
\usepackage{epsfig}
\usepackage{graphicx}
\usepackage{subfigure}
\usepackage{amsmath}
\usepackage{amssymb}
\usepackage{hyperref}

\usepackage{mathrsfs}
\usepackage{algorithm}
\usepackage{algorithmic}
\usepackage{array}
\usepackage{booktabs}
\usepackage{multirow}

\newcommand{\INPUT}{\item[\myinput]}
\newcommand{\myinput}{\textbf{Preparation:}}

\newcommand{\MYWHILE}{\item[\mywhile]}
\newcommand{\mywhile}{\textbf{repeat}}

\newcommand{\MYENDWHILE}{\item[\myendwhile]}
\newcommand{\myendwhile}{\textbf{until}}
\ifCLASSINFOpdf
\else
\fi

\begin{document}

%
\title{Bit-Scalable Deep Hashing with Regularized Similarity Learning for Image Retrieval and Person Re-identification}
%
%
%

\author{Ruimao~Zhang,
        Liang~Lin,
        Rui~Zhang,
        Wangmeng~Zuo,
        and~Lei~Zhang
\IEEEcompsocitemizethanks{This work was supported in part by the Hong Kong
Scholar program, in part by the Guangdong Natural Science Foundation under Grant S2013050014548 and Grant 2014A030313201, and in part by the Program of Guangzhou Zhujiang Star of Science and Technology under Grant 2013J2200067.

\IEEEcompsocthanksitem R. Zhang, L. Lin and R. Zhang are with Sun Yat-sen University, Guangzhou 510006, China. L. Lin is also with the Department of Computing, the Hong Kong Polytechnic University, Kowloon, Hong Kong, China. E-mail: r.m.zhang1989@gmail.com; linliang@ieee.org; rayz0620@gmail.com}
\IEEEcompsocitemizethanks{\IEEEcompsocthanksitem W. Zuo is with School of Computer Science and Technology, Harbin Institute of Technology, Harbin 150001, China. E-mail: cswmzuo@gmail.com}
\IEEEcompsocitemizethanks{\IEEEcompsocthanksitem L. Zhang is with the Department of Computing, the Hong Kong Polytechnic University, Kowloon, Hong Kong, China. E-mail: cslzhang@comp.polyu.edu.hk}
}

\markboth{IEEE Transactions on Image Processing}%
{Zhang \MakeLowercase{\textit{et al.}}: Bit-Scalable Deep Hashing with Regularized Similarity Learning for Image Retrieval}
%



\maketitle

\begin{abstract}

Extracting informative image features and learning effective approximate hashing functions are two crucial steps in image retrieval . Conventional methods often study these two steps separately, {\em e.g.}, learning hash functions from a predefined hand-crafted feature space. Meanwhile, the bit lengths of output hashing codes are preset in most previous methods, neglecting the significance level of different bits and restricting their practical flexibility. To address these issues, we propose a supervised learning framework to generate compact and bit-scalable hashing codes directly from raw images. We pose hashing learning as a problem of regularized similarity learning. Specifically, we organize the training images into a batch of triplet samples, each sample containing two images with the same label and one with a different label. With these triplet samples, we maximize the margin between matched pairs and mismatched pairs in the Hamming space. In addition, a regularization term is introduced to enforce the adjacency consistency, {\em i.e.}, images of similar appearances should have similar codes. The deep convolutional neural network is utilized to train the model in an end-to-end fashion, where discriminative image features and hash functions are simultaneously optimized. Furthermore, each bit of our hashing codes is unequally weighted so that we can manipulate the code lengths by truncating the insignificant bits. Our framework outperforms state-of-the-arts on public benchmarks of similar image search and also achieves promising results in the application of person re-identification in surveillance. It is also shown that the generated bit-scalable hashing codes well preserve the discriminative powers with shorter code lengths.

\end{abstract}


%
\IEEEpeerreviewmaketitle

\section{Introduction}\label{sec:introduction}

With the fast growth of image or video collections, hashing techniques have been receiving increasing attentions in large scale image retrieval~\cite{HammingEmbedding}\cite{FineGrainedSimilarity}\cite{DBLP:ContextualHashing}\cite{DBLP:VideoHash} and related applications ({\em e.g.} person re-identification in surveillance). Recently, many learning-based hashing schemes have been proposed \cite{DBLP:ColumGenerHash}\cite{DBLP:IsotropicHash}\cite{DBLP:KSH}\cite{DBLP:Sparse-EmbeddingHashing-TIP}, which target on learning a compact and similarity-preserving representation such that similar images are mapped to nearby binary hash codes in the Hamming space. Among them, the supervised approaches~\cite{DBLP:KSH}\cite{SEMI-SUPERVISED} have shown great potentials by exploiting the supervised information (\textit{e.g.}, class labels) in hashing learning.

Traditional image retrieval systems based on supervised hashing learning usually involve two crucial steps. First, the stored images are encoded with a vector of hand-crafted descriptors in order to capture the image semantics against image noises and other redundant information. Second, the hashing learning is posed as either a pointwise or a pairwise optimization \cite{DBLP:journals/jmlr/metriclearning1}\cite{LDF} problem to preserve the pointwise or pairwise label information in the learned Hamming space. However, the above two steps are mostly studied as two independent problems, which leads to unsatisfying results. The feature representation may not be tailored to the objective of hashing learning. Moreover, the hand-crafted feature engineering often requires much domain knowledge and heavy tuning.

On the other hand, most existing hashing learning approaches generate the hashing codes with preset lengths ({\em e.g.}, 16, 32 or 64 bits)~\cite{DBLP:ColumGenerHash}\cite{DBLP:KSH}\cite{DBLP:MLH}, but one often requires hashing codes of different lengths under different scenarios. For example, the shorter codes are beneficial to devices with limited computation resources ({\em e.g.}, mobile devices), while longer codes are used for pursuing higher accuracy. To cope with such requirements, one conventional solution is to store several versions of hashing codes in different bit lengths, consequently causing extra computation and storage.  In literature, several bit-scalable hashing methods are exploited. They usually generate hashing codes bit by bit in a significance descent way, \textit{i.e.,} the former bits are learned typically more significant than latter, so that one can simply pick desired number of bits from the top of the hashing codes~\cite{DBLP:SH}\cite{DBLP:IsotropicHash}\cite{DBLP:ITQ}.  However, these methods usually require to carefully design the embedded feature space and their performances may dramatically fall when shortening the hashing codes.



A novel supervised {\bf Bit-Scalable Deep Hashing} framework\footnote{Source code available at: \url{http://vision.sysu.edu.cn/projects/DeepHashing/}} is proposed in this work to address the above mentioned issues, and we validate its effectiveness on the tasks of general image retrieval and person re-identification across disjoint camera views. The convolutional neural network (CNN) is utilized to build the end-to-end relation between the raw image data and the binary hashing codes for fast indexing. Moreover, each bit of these output hashing codes is weighted according to their significance so that we can manipulate the code lengths by truncating the insignificant bits. The hashing codes of arbitrary lengths (less than the original codes) can then be easily obtained without extra computation. In the following, we overview the main components of our framework and summarize the advantages.

\begin{figure}
\includegraphics[width=3.2in]{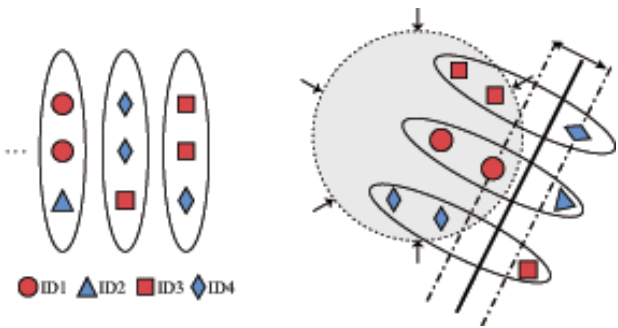}
\centering
\vspace{0.5em}
\caption{Illustration of the triplet-based regularized similarity learning. A batch of triplet samples (represented by the solid eclipses) are organized. Each triplet contains three images (represented by dots with different shapes) with only two of them having the same label. The margin between the matched pairs and the mismatched pairs are maximized in the Hamming space, while regularization (indicated by the gray dashed circle) is incorporated to constrain the images of similar appearances to have similar hashing codes.}
\label{fig:sketch}
\end{figure}

(I). We present a novel formulation of relative similarity comparison based on the triplet-based model. As discussed in \cite{DBLP:journals/jmlr/metriclearning1}\cite{FineGrainedSimilarity}\cite{DBLP:DingShengYong}, the triplet-like samples can well capture the intra-class and inter-class variations in the ranking optimization. In hashing learning, however, the images of similar appearances are also expected to have close hashing codes in the Hamming space. Therefore, we extend the triplet-based relative comparison by incorporating a regularization term, partially motivated by the recently proposed Laplacian Sparse Coding \cite{DBLP:LaplacianSparseCoding}. Fig.~\ref{fig:sketch} illustrates our formulation. Specifically, we organize training images into a large number of triplet samples, and each sample contains three images with only two of them having the same label. Then, for each triplet sample, we formulate the hashing learning as a joint task of maximizing the relative distance between the matched pair and the mismatched pair, while preserving the adjacency relation of images in the Hamming space.

%

(II). We adopt the deep CNN architecture to extract the discriminative features from the input images, where the convolutional layers, max-pooling operators, and one full connection layer are stacked up. Over the features generated by previous layers, we impose one fully-connected layer and one tanh-like layer to output the binary hashing codes. On the top of our model, an element-wise layer is designed to weigh each bin of the hashing codes for bit-scalable hashing. In our deep model, the hash function learning and the feature learning are jointly optimized via backward propagation. Moreover, the generated bit-scalable hash codes are able to well preserve the matching accuracy with varying code lengths.

(III). To cope with the large amount of stored images, we implement our learning algorithm in a batch-process fashion. In each round of learning, we first organize the triplet samples from a randomly selected subset (\textit{i.e.}, $150\sim 200$) of the training images, and then utilize the stochastic gradient descent (SGD) method for parameter learning. Since one image can be included in several triplet samples, we calculate the partial derivative on images instead of on triplet samples. The computational cost is thus much reduced and it is linear to the selected subset of images.

This paper makes three main contributions to image retrieval. i) First, it unifies feature learning and hash function learning via deep neural networks, and the proposed bit-scalable hashing learning can effectively improves the flexibility of image retrieval. ii) Second, it presents a novel formulation (\textit{i.e.}, the regularized triplet-based comparison) for hashing learning, and it is general to be extended to other similar tasks. iii) Third, our extensive experiments on standard benchmarks demonstrate that the learned hashing codes well preserve the instance-level similarity and outperforms state-of-the-art hashing learning approaches. Moreover, we successfully apply our hashing method to the application of {\bf person re-identification} in surveillance. This task, aiming at retrieving the same individual across several non-overlapped cameras, has received increasingly attention in computer vision research.

The rest of the paper is organized as follows. Section~\ref{sec:RelatedWork} presents a brief review of related work.  Section~\ref{sec:DRPCH} introduces our hashing learning framework, followed by a discussion of learning algorithm in Section~\ref{sec:Optimization}. The experimental results, comparisons and component analysis are presented in Section~\ref{sec:Experiment}. Section~\ref{sec:conclusion} concludes the paper.

\section{Related Work}
\label{sec:RelatedWork}

Recently, hashing is becoming an important technique for fast approximate similarity search. Generally speaking, hashing methods can be categorized into two classes: data-independent and data-dependent. Data-independent methods randomly generate a set of hash functions without any training, and they usually make the hashing codes scattered to keep the matching accuracy~\cite{DBLP:MIN-HASH-2}. Exemplars include Locality Sensitive Hashing \cite{DBLP:LSH} and its variants \cite{DBLP:kernel-LSH}, and the Min-Hash algorithms~\cite{DBLP:MIN-HASH}.

On the other hand, data-dependent hashing methods focus on how to learn  compact hashing codes from the training data. These learning-based approaches usually comprise two stages: i) projecting the high dimensional features onto the lower dimensional space, and ii) quantizing the generated real-valued representations into binary codes. Specifically, unsupervised methods learn the hash functions using unlabeled data, which seek to propagate neighborhood relation of samples from a certain metric space into the Hamming space~\cite{DBLP:SH}\cite{DBLP:SparseSpectralHashing}\cite{DBLP:Self-TaughtHashing}\cite{DBLP:GraphHash}\cite{DBLP:LLH}. For example, Spectral Hashing \cite{DBLP:SH} constructs the global graph with $L_2$ distance and optimizes the graph Laplacian cost function in the Hamming space. Locally Linear Hash \cite{DBLP:LLH} pursues the structures of manifolds in the Hamming space and optimizes such structures by locality-sensitive sparse coding. For the semi-supervised~\cite{DBLP:SequentialProjection}\cite{DBLP:Semi-SupervisedHashing-PAMI} and supervised methods~\cite{DBLP:MLH}\cite{DBLP:OnlineHash}\cite{DBLP:ColumGenerHash}\cite{DBLP:HammingDistance}\cite{DBLP:KSH}, richer similarity information of training samples (\textit{e.g.}, pairwise similarity or relative distance comparison \cite{DBLP:HammingDistance}) is exploited to improve the hashing learning. For example, Wang \textit{et al.}~\cite{DBLP:Semi-SupervisedHashing-PAMI} proposed a semi-supervised hashing framework, which minimizes the empirical error on the labeled data while maximizing the variance over labeled and unlabeled data simultaneously.  Norouzi \textit{et al.} introduced the Minimal Loss Hashing \cite{DBLP:MLH} based on structured prediction with latent variables and a hinge-like loss function. Following \cite{DBLP:MLH}, Huang \textit{et al.} proposed the Online Hashing \cite{DBLP:OnlineHash} to update the hash function incrementally. Column Generation Hashing \cite{DBLP:ColumGenerHash} aims to learn hash function based on proximity comparison information and preserve the data relationship based on large-margin principle. In \cite{DBLP:HammingDistance}, Norouzi \textit{et al.} also employed triplet-based model with loss-augmented inference and showed very good results in image retrieval and classification.  However, in each iteration, the time cost of such structured prediction method heavily depends on the scale of data and the length of hash code. Liu \textit{et al.} proposed the Kernel-based Supervised Hashing \cite{DBLP:KSH}, in which the non-linear kernel was utilized with triplet-based hash function learning.

Rather than using hand-crafted representations \cite{PISA2015}, extracting features and capturing contextual relations with deep learning techniques have shown great potential in various vision recognition tasks such as image classification and object detection~\cite{ImageNetClassification}\cite{lin2015discriminatively}\cite{3D-Convolutional}\cite{DBLP:RCNN}\cite{joint-task}. Very recently, Wu \textit{et al.} \cite{FineGrainedSimilarity} proposed a learning-to-rank framework based on multi-scale neural networks, and showed promising performance on capturing fine-grained image similarity. Pre-training on the large-scale image classification database ({\em i.e.}, ImageNet~\cite{ImageNetClassification}) was used in this model. Another related work was proposed by Xia \textit{et al.} \cite{DBLP:Panyan-AAAI14}, which utilizes CNN for supervised hashing learning. They first produced the hashing codes of images by decomposing the pairwise similarity matrix, and then learned the mapping functions from images to the codes. This method, however, may fail to deal with large-scale data due to the matrix decomposition operation. Our approach proposed in this paper  advances the above methods in the novel regularized triplet-based formulation and the bit-scalable hashing generation.

\section{Bit-Scalable Deep Hashing Framework}
\label{sec:DRPCH}

\begin{figure*}[ht]
\includegraphics[width=6.5in]{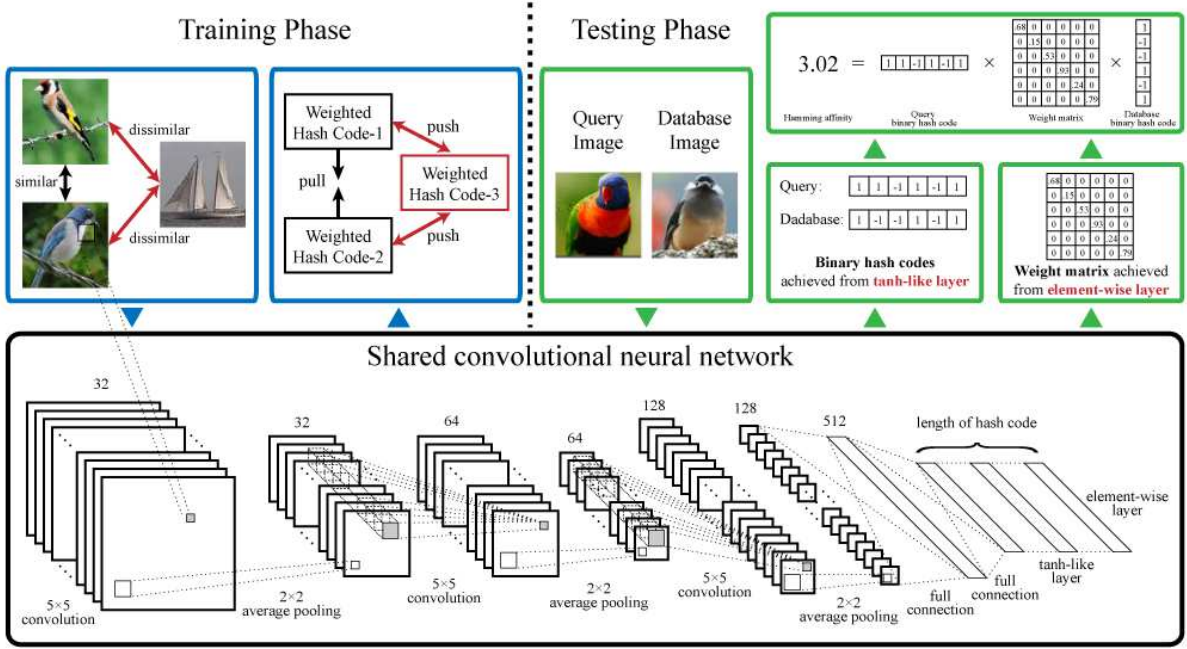}
\caption{The bit-scalable deep hashing learning framework. The bottom panel shows the deep architecture of neural network that produces the hashing code with the weight matrix by taking raw images as inputs. The training stage is illustrated in the left up panel, where we train the network with triplet-based similarity learning. An example of hashing retrieval is presented in the right up panel, where the similarity is measured by the Hamming affinity.}
\label{fig:framework}
\end{figure*}

The objective of hashing learning is to seek the mapping function $h(x)$ that projects $p$-dimensional real valued feature vector $x\in R^p$ onto $q$-dimensional binary hash code $h\in \{-1,1\}^q$, while preserving semantic consistency of each pair. In this section we introduce our bit-scalabe deep hashing framework, which is illustrated in Fig. \ref{fig:framework}. Instead of learning hash function on hand-crafted feature space, we integrate image feature learning and hashing learning into a nonlinear transformation function $\phi(\cdot)$ taking the raw image as input. In addition, we introduce a weight vector $\textbf{w} = [w_1, ..., w_q]^T$ to weigh each bit of the output hash codes, which represents the significance of each bit in measuring similarity. In our framework, a deep architecture of CNNs is developed to jointly learn $\phi(\cdot)$ and $\textbf{w}$.

We express the nonlinear hash function as a parametric form:
\begin{equation}\label{eq-1}
h = sign(\phi(I))
\end{equation}
where $sign(\cdot)$ denotes the element wise sign function, $I$ is a raw image. Different from our model, many state-of-the-art methods are designed to learn a hash function $sign(A^T x)$ of linear projection $A^T x$, where $x$ is a hand-crafted feature representation. With the weight $\bf{w}$, we employ the weighted Hamming affinity~\cite{MultidimensionalSpectralHashing} to measure the dissimilarity between two hashing codes, which is expressed as a linear combination of the agreement between the two codes:
\begin{equation}\label{eq_HamAff}
\mathcal{H} (h(x_j), h(x_k)) = h(x_j) \widetilde{\textbf{w}} h(x_k) = -\sum_i w_i^2 h_i(x_j) h_i(x_k)
\end{equation}
where $\widetilde{\textbf{w}}$ is the diagonal matrix whose diagonal value is represented as $\widetilde{\textbf{w}}(i,i) = w_i^2$.


The weighted hash code brings several distinctive advantages in hash learning. (i) Instead of treating each bit equally, we can produce more effective hashing code by assigning different weights to different bits. (ii) By truncating the insignificant bins corresponding to small weights, we can flexibly manipulate the code lengths for different scenarios (e.g., adapting to computational resources). (iii) The weighted Hamming distance can be naturally degenerated into the conventional version.


\subsection{Formulation}

We organize the training images into triplet samples, and pose the hashing learning problem as a problem of regularized similarity learning. Each triplet contains three images with only two of them having the same label and the other one having a different label. We define a Max-Margin term embedded in the Hamming space to maximize the margin between the matched pairs and the mismatched pairs, which is similar to the fine-grained image similarity model in~\cite{FineGrainedSimilarity}. Intuitively, this term guarantees the learned hashing codes to preserve the ranking orders of images according to the annotated semantics.

%
%

Let  $\mathcal{D}=\{(I_i,I_i^+,I_i^-)\}_{i=1}^N$ be a set of triplet units, in which $I_i$ and $I_i^+$ are two images having the same label, $I_i$ and $I_i^-$ are two mismatched images, and $N$ is the total number of training triplets. Let $\omega$ denote the parameters of hashing functions and $h(I_i) \in \{-1,1\}^q$ denote the $q$ bits hashing code of image $I_i$. For simplicity, we use $h_i$ to replace $h(I_i)$, and use $h_i^+$ and $h_i^-$ to denote $h(I_i^+)$ and $h(I_i^-)$, respectively. With the triplet-based samples, the loss function of the Max-Margin term can be  written as:

\begin{equation}\label{eq-2}
\min \sum_{i,i^+,i^-} {\Phi_{\textbf{w}}(h_i,h_i^+,h_i^-)}
\end{equation}
where $\Phi_{\textbf{w}}(\cdot,\cdot,\cdot)$ is the max-margin loss defined for one triplet. We require that the weighted Hamming affinity should satisfy the following constraint:
\begin{equation}\label{eq-3}
\mathcal{H} (h_i, h_i^+) < \mathcal{H} (h_i, h_i^-)
\end{equation}
Then, we have the following hinge-like loss function:

\begin{equation}\label{eq_MaxMarTerm}
\sum_{i,i^+,i^-} \Phi_\textbf{w}(h_i,h_i^+,h_i^-) = \sum_{i=1}^N \max\{G_\textbf{w}(h_i,h_i^+,h_i^-),C\}
\end{equation}
where $G(h_i,h_i^+,h_i^-) = \mathcal{H} (h_i, h_i^+) - \mathcal{H} (h_i, h_i^-)$, and $\mathcal{H} (\cdot, \cdot)$ is defined in Eq.~(\ref{eq_HamAff}). The max operator and constant $C$ are introduced to enhance the robustness again outliers, as defined in SVMs. We set \textbf{$C=-q/2$} throughout the experiments.

%

In addition to preserving the image ranking, we also encourage the adjacency relation of images in the original appearance space to be stressed with the learned hashing codes. Thus, we define the following regularization term:

\begin{equation}\label{eq_regTerm}
\sum_{i,j} \Psi_\textbf{w}(h_i,h_j) = \frac{1}{2}\sum_{ij}\mathcal{H} (h_i, h_j) S_{ij}
\end{equation}
where $S_{ij}$ represents the similarity between an image pair $(I_i,I_j)$ over the training set. As introduced in \cite{DBLP:LaplacianSparseCoding}, $S_{ij}$ is large when two images are similar and small when they are dissimilar. The way of specifying $S_{ij}$ will be discussed in Sec.~\ref{sec:Experiment}. Following \cite{DBLP:LaplacianSparseCoding}, we define the diagonal degree matrix $U$ with $U_{ii}=\sum_{j}S_{ij}$. The Laplacian matrix \cite{DBLP:LaplacianEigenmap} can then be defined as $L=U-S$ \cite{DBLP:LaplacianSparseCoding}, and we can rewrite the regularization term Eq.~(\ref{eq_regTerm}) into the following form:
\begin{equation}\label{eq_regTerm2}
\sum_{i,j} \Psi_\textbf{w}(h_i,h_j) = \frac{1}{2} \mathrm{tr}(HLH^T)
\end{equation}
where $H=[h_1 {\widetilde{\textbf{w}}}^{\frac{1}{2}},h_2 {\widetilde{\textbf{w}}}^{\frac{1}{2}},...,h_M {\widetilde{\textbf{w}}}^{\frac{1}{2}}]$ and $M$ is the total number of images utilized to generate $\mathcal{D}$, and $\mathrm{tr}(\cdot)$ denotes the trace operator.

By combining Eq.(\ref{eq_MaxMarTerm}) and Eq.(\ref{eq_regTerm2}), we have the following regularized triplet-based comparison model:

\begin{equation}\label{eq_PriModel}
\min_{\textbf{w}, \omega}\sum_{t=1}^N \max\{G_\textbf{w}(h_i,h_i^+,h_i^-),C\} + \lambda\mathrm{tr}(HLH^T)
\end{equation}
Since the hash codes are binary, the above objective is discontinuous and
nondifferentiable and thus is difficult to be optimized via gradient descent. To address this problem, we propose a tanh-like approximation $o(v)$ of the sign function:

\begin{equation}\label{eq-9}
o(v)=\frac {1 - e ^ {-\beta v} }  {1 + e ^ {-\beta v}}
\end{equation}
where $\beta$ is a tuning parameter to control the smoothness. When $\beta=2$, Eq.~(\ref{eq-9}) is a standard hyperbolic tangent function. When  $\beta$ is very large, the activation function in Eq.~(\ref{eq-9}) approximates to a sign function. In this paper, $\beta$ is increasing from $2$ to $1000$ in the iterations of learning. In the test stage, the sign function is adopted as the activation function to obtain the discrete hash code.

With $o(v)$, the hash code $h_i$ can be approximated by $r_i \in[-1,1]^q$:

\begin{equation}\label{ApproximateHash}
r = o(\phi(I))
\end{equation}
We further define $D_\textbf{w}(r_i,r_i^+,r_i^-)$ to approximate $G_\textbf{w}(h_i,h_i^+,h_i^-)$ as follows:

\begin{equation}\label{ApproximateG}
D_\textbf{w}(r_i,r_i^+,r_i^-) = \mathcal{M} (r_i, r_i^+) - \mathcal{M} (r_i, r_i^-)
\end{equation}
where $\mathcal{M}(\cdot,\cdot)$ is the weighted Euclidean distance between the approximated hash codes:

\begin{equation}\label{Eculidean}
\mathcal{M} (r_i,r_j) = \|r_i{\widetilde{\textbf{w}}}^{\frac{1}{2}}- r_j{\widetilde{\textbf{w}}}^{\frac{1}{2}}\|_2^2
\end{equation}

Finally, the continuous approximation of the regularized triplet-based learning model is written as:

\begin{equation}\label{eq_Lapmodel}
\min_{\textbf{w}, \omega}\sum_{i=1}^N \max\{D_\textbf{w}(r_i,r_i^+,r_i^-),C\} +\lambda \mathrm{tr}(RLR^T)
\end{equation}
where $R=[r_1 {\widetilde{\textbf{w}}}^{\frac{1}{2}},r_2 {\widetilde{\textbf{w}}}^{\frac{1}{2}},...,r_M {\widetilde{\textbf{w}}}^{\frac{1}{2}}]$.

An obvious advantage of binary hashing is that bit-wise XOR or lookup table can be adopted to measure the distances between hash codes. Even the proposed weighted hash makes it impossible to use this efficient searching strategy, we develop a lookup table (LUT) based approach to rapidly return the weighted affinity between hash codes. For simplicity, let $l$ denotes the length of hash code. We can set up a lookup table with the length $2^l$, which equals to the total number of candidate XOR results between two hash codes. Because the hash weights are pre-trained and fixed in the searching stage, the weighted hamming affinity of each XOR result can be calculated in advance and stored in the lookup table as the item. In this way, the ranking list can be efficiently returned by the table lookup search. Although this method provides a feasible solution for the efficient searching, the storage of the table is exploding as $l$ becomes large. A reasonable strategy to handle this point is to split the hash code into different parts with equal length (set as 8 in this paper). Each part is associated with a special sub-table with fixed length. The output of each sub-table is the weighted similarity value of the corresponding part. The overall hash affinity can be calculated by accumulating the weighted similarity values from all parts, and then the final ranking list is generated based on the overall hash affinity.

\subsection{Deep Architecture}
\label{Arch}

In order to incorporate the feature representation learning and binary hash code learning into an end-to-end learning framework, we introduce the deep  CNN into our hash learning process. Fig.~\ref{fig:framework} shows the overall network architecture, which consists of ten layers. The first six layers form the convolution-pooling network with rectified linear activation and average pooling operation. We use $32$, $64$, and $128$ filters with size $5 \times 5$ in the first, second and third convolutional layers and the stride is 2 pixels in every convolution layer. The stride for pooling is 1 and we set the pooling operator size as $2\times 2$. The last four layers include two standard fully connected layers, a tangent like layer to output hash codes, and an element-wise connected layer to weigh each bit of hash code. The number of units is $512$ in the first fully-connected layer and the output of the second fully-connected layer equals to the length of hash code. The activation function of the second fully-connected layer is the tanh-like function defined in Eq.~(\ref{eq-9}), and rectified linear activation function is adopted for the other layers.

\section{Learning Algorithm}
\label{sec:Optimization}
In this section, we present how to optimize the network parameters given a  set of training images and a fixed number of triplets.  The implementation details about generating triplets from labeled images and training the network with batch mode are also presented at the end of this section.

\subsection{Joint Optimization}

Let's first consider the learning algorithm with the loss function defined in Eq.(\ref{eq_Lapmodel}). The parameter optimization of varied length hashing learning is the same. For simplicity, we consider the parameters in the network as a whole and define $\varpi=[\omega,\textbf{w}]$. Thus, the loss function can be expressed as:

\begin{equation}\label{eq-12}
\mathcal{L}(\varpi)=\sum_{i=1}^N \max\{D_\textbf{w}(r_i,r_i^+,r_i^-),C\} +\lambda \mathrm{tr}(RLR^T)
\end{equation}

In order to employ back propagation algorithm to optimize the network parameters, we compute the partial derivative of the objective function:

\begin{equation}\label{eq-13}
\frac{\partial \mathcal{L}}{\partial \varpi_k} = \sum_{i=1}^N d_\textbf{w}(r_i,r_i^+,r_i^-) + \lambda\sum_{j=1}^M f_\textbf{w}(r_j)
\end{equation}
By the definition of $D_\textbf{w}(r_i,r_i^+,r_i^-)$ in Eq.(\ref{eq_Lapmodel}), we obtain the gradient as follows:

\begin{equation}\label{eq-14}
 d_\textbf{w}(r_i,r_i^+,r_i^-) =
  \left\{
   \begin{array}{rl}
   \frac{\partial D_\textbf{w}(r_i,r_i^+,r_i^-)}{\partial \varpi_k}& \mbox{,  if $D_\textbf{w}(r_i,r_i^+,r_i^-)>C$}\\
   0 \quad\quad&\mbox{,  if $D_\textbf{w}(r_i,r_i^+,r_i^-)\leq C$}\\
   \end{array}
  \right.
\end{equation}

\begin{equation}\label{eq-15}
\begin{split}
\frac{\partial D_\textbf{w}(r_i,r_i^+,r_i^-)}{\partial \varpi_k} &= 2 ( r_i{\widetilde{\textbf{w}}}^{\frac{1}{2}}-r_i^+{\widetilde{\textbf{w}}}^{\frac{1}{2}} )^{'} \cdot \frac{\partial (r_i{\widetilde{\textbf{w}}}^{\frac{1}{2}})- \partial (r_i^+{\widetilde{\textbf{w}}}^{\frac{1}{2}})}{\partial \varpi_k} \\
&-2 ( r_i{\widetilde{\textbf{w}}}^{\frac{1}{2}}-r_i^- {\widetilde{\textbf{w}}}^{\frac{1}{2}})^{'} \cdot \frac{\partial (r_i{\widetilde{\textbf{w}}}^{\frac{1}{2}})- \partial (r_i^-{\widetilde{\textbf{w}}}^{\frac{1}{2}})}{\partial \varpi_k}
\end{split}
\end{equation}
It is clear that the gradient of each triplet can be calculated by the value of $(r_j{\widetilde{\textbf{w}}}^{\frac{1}{2}})$ and $\frac{\partial (r_j{\widetilde{\textbf{w}}}^{\frac{1}{2}})}{\partial \varpi_k}$ for a single image. Thus, the gradient of the first term in Eq.(\ref{eq_Lapmodel}) can be obtained by the forward and backward propagation  for each image in the triplet.

On the other hand, we can rewrite the optimization of the second term in Eq.(\ref{eq_Lapmodel}) with respect to $r_j$ as follows:

\begin{equation}\label{eq-16}
\begin{split}
\mathrm{tr}(RLR^T) &= (r_j{\widetilde{\textbf{w}}}^{\frac{1}{2}})^T(RL_j) + (RL_j)^T(r_j{\widetilde{\textbf{w}}}^{\frac{1}{2}}) \\
 &- (r_j{\widetilde{\textbf{w}}}^{\frac{1}{2}})^TL_{ii}(r_j{\widetilde{\textbf{w}}}^{\frac{1}{2}})
\end{split}
\end{equation}
where $L_j$ is the $j$-th column of $L$. Following \cite{DBLP:LaplacianSparseCoding}, we define the matrix $R_{-j}$ as the submatrix formed by removing the $j$-th column of matrix $R$, and define  the vector $L_{j,-j}$ as the subvector after removing the $j$-th entry of vector $L_j$. Then $f(r_j)$ in Eq.(\ref{eq-13}) can be calculated by

\begin{equation}\label{eq-17}
f_\textbf{w}(r_j)= ( R_{-j}L_{j,-j} + L_{jj}(r_j{\widetilde{\textbf{w}}}^{\frac{1}{2}}) )\cdot  \frac{\partial (r_j{\widetilde{\textbf{w}}}^{\frac{1}{2}})}{\partial \varpi_k}
\end{equation}

We can observe that the gradient of the second term in Eq.(\ref{eq_Lapmodel}) can also be computed through $(r_j{\widetilde{\textbf{w}}}^{\frac{1}{2}})$ and $\frac{\partial (r_j{\widetilde{\textbf{w}}}^{\frac{1}{2}})}{\partial \varpi_k}$. Reviewing the discussions above, the overall process of joint optimization is summarized as follows: (1) calculating $(r_j{\widetilde{\textbf{w}}}^{\frac{1}{2}})$ for a certain image $I_j$ by forward propagation; (2) calculating $\frac{\partial (r_j{\widetilde{\textbf{w}}}^{\frac{1}{2}})}{\partial \varpi_k}$ by backward propagation; (3) calculating each $\frac{\partial D_\textbf{w}(r_j,r_j^+,r_j^-)}{\partial \varpi_k}$ corresponding to $I_j$ by Eq.(\ref{eq-15}); (4) summing the gradient $\frac{\partial \mathcal{L}}{\partial \varpi_k}$ according to Eq.(\ref{eq-13}).

\subsection{Acceleration}

In the above discussed optimization, both the first and second terms of loss function need to know $(r_j{\widetilde{\textbf{w}}}^{\frac{1}{2}})$ and $\frac{\partial (r_j{\widetilde{\textbf{w}}}^{\frac{1}{2}})}{\partial \varpi_k}$ to calculate the partial derivative. The only difference is that  the first term needs to compute triplet based gradient according to Eq.(\ref{eq-15}), but the second term does not. Discovering this difference inspires us to look for a more effective optimization algorithm which depends only on image based gradient.

We observe that the overall gradient can in fact be obtained from gradient calculated for each image separately. We first consider the second term of Eq.(\ref{eq-12}), whose partial derivative depends on a single image. In contrast, it is difficult to write the first term of Eq.(\ref{eq-12}) directly as the sum of the cost on images, which takes the following form:

\begin{equation}\label{eq-18}
\mathcal{L}(\varpi)=\frac{1}{N} \sum_{i=1}^N \mathcal{J}((r_i{\widetilde{\textbf{w}}}^{\frac{1}{2}}),(r_i^+{\widetilde{\textbf{w}}}^{\frac{1}{2}}),(r_i^-{\widetilde{\textbf{w}}}^{\frac{1}{2}}))
\end{equation}
where $N$ is the total number of triplets. Fortunately, because the loss function for a specific triplet is defined by the outputs of the images in this triplet, the total loss can also be considered as follows:

\begin{equation}\label{eq-19}
\mathcal{L}(\varpi)=\mathcal{L}( (r_1{\widetilde{\textbf{w}}}^{\frac{1}{2}}),(r_2{\widetilde{\textbf{w}}}^{\frac{1}{2}}),...(r_j{\widetilde{\textbf{w}}}^{\frac{1}{2}}),..,(r_M{\widetilde{\textbf{w}}}^{\frac{1}{2}}) )
\end{equation}
where $r_j$ corresponds to  the distinct image in some triplets. $M$ indicates the total number of images adopted in triplet set $\mathcal{D}$. The derivative rule gives us the following equation:

\begin{equation}\label{eq-20}
\frac{\partial \mathcal{L}}{\partial \varpi} = \sum_{i=1}^N \frac{\partial \mathcal{L}}{\partial (r_i{\widetilde{\textbf{w}}}^{\frac{1}{2}})} \frac{\partial (r_i{\widetilde{\textbf{w}}}^{\frac{1}{2}})}{\partial \varpi}
\end{equation}

Eq.(\ref{eq-20}) is very similar to traditional image based partial derivative. The only variation is the way in which the partial differential is calculated with respect to the image outputs. In the traditional image based loss function, this calculation depends on only one image, whereas in the triplet-based loss function, it depends on the outputs of all images in the triplets. Algorithm~\ref{alg1} provides the sketch of our hashing learning framework and Algorithm~\ref{alg2} presents how to compute the partial differential with respect to the network output. Such an image-based gradient calculation method effectively reduces the computational cost, which is significant for handling large scale data.

\begin{small}
\begin{algorithm}[ht]
\caption{Deep hashing learning}
\label{alg1}
\begin{algorithmic}
\REQUIRE ~~\\
   Training triplets $\mathcal{D}$.
\ENSURE ~~\\                           
    The network parameters $\omega$.
\INPUT ~~\\
    Collect all the distinct images $\{I_j\}$ in $\mathcal{D}$.
\MYWHILE
    \STATE
    \begin{itemize}
  \setlength{\itemsep}{1pt}
  \setlength{\parskip}{3pt}
  \setlength{\parsep}{10pt}
  \item[1.] Calculate outputs $(r_j{\widetilde{\textbf{w}}}^{\frac{1}{2}})$ of image $I_j$ by forward propagation.

  \textbf{repeat}
  \setlength{\parskip}{-1pt}
     \begin{itemize}
     \setlength{\itemsep}{1pt}
     \setlength{\parskip}{0pt}
     \setlength{\parsep}{10pt}
       \item[a)] Calculate $\frac{\partial \mathcal{L}}{\partial (r_j{\widetilde{\textbf{w}}}^{\frac{1}{2}})}$ for image $I_j$ by Algorithm ~\ref{alg2};
       \item[b)] Calculate $\frac{\partial \mathcal{L}}{\partial \varpi_k}(r_j{\widetilde{\textbf{w}}}^{\frac{1}{2}})$ utilizing back propagation;
       \item[b)] Sum the partial derivative: $\frac{\partial \mathcal{L}}{\partial \varpi_k} +=\frac{\partial \mathcal{L}}{\partial \varpi_k}(r_j{\widetilde{\textbf{w}}}^{\frac{1}{2}})$;
     \end{itemize}

  \textbf{until} Traverse all the images in $\{I_j\}$;

  \setlength{\parskip}{0pt}
  \item[2.] Update $\varpi_k^t=\varpi_k^{t-1}-\psi_t \frac{\partial \mathcal{L}}{\partial \varpi_k}$ and $t \leftarrow t+1$.

      \end{itemize}
\MYENDWHILE
$t>T$.
\end{algorithmic}
\end{algorithm}
\end{small}

\begin{small}
\begin{algorithm}[ht]
\caption{Image Based Partial Derivative}
\label{alg2}
\begin{algorithmic}[1]
\REQUIRE ~~\\
    Training triplet set $\mathcal{D}$, image $I_j$, matrix $D$ in Eq.(\ref{eq_Lapmodel}).
\ENSURE ~~\\                           
    The partial derivative of $\frac{\partial \mathcal{L}}{\partial (r_j{\widetilde{\textbf{w}}}^{\frac{1}{2}})}$.
\INPUT ~~\\
    $pSum = 0$;

\STATE
\textbf{for all } $(I_i,I_i^+,I_i^-)$ \textbf{do }
\STATE
\ \ \ \textbf{if} $I_j=I_i$ \textbf{then}
\STATE
\ \ \ \ \ \  $pSum += 2(r_i^-{\widetilde{\textbf{w}}}^{\frac{1}{2}} - r_i^+{\widetilde{\textbf{w}}}^{\frac{1}{2}})$
\STATE
\ \ \  \textbf{else if} $I_j=I_i^+$ \textbf{then}
\STATE
\ \ \ \ \ \  $pSum -= 2(r_i{\widetilde{\textbf{w}}}^{\frac{1}{2}} - r_i^+{\widetilde{\textbf{w}}}^{\frac{1}{2}})$
\STATE
\ \ \  \textbf{else if} $I_j=I_i^-$ \textbf{then}
\STATE
\ \ \ \ \ \  $pSum += 2(r_i{\widetilde{\textbf{w}}}^{\frac{1}{2}} - r_i^-{\widetilde{\textbf{w}}}^{\frac{1}{2}})$
\STATE
\ \ \  \textbf{end if}
\STATE
\textbf{end for}
\STATE
Calculate $f_\textbf{w}(r_j)$ according to Eq.(\ref{eq-17}).
\STATE
Return $\frac{\partial \mathcal{L}}{\partial (r_j{\widetilde{\textbf{w}}}^{\frac{1}{2}})} = pSum + \lambda f_\textbf{w}(r_j)$.
\end{algorithmic}
\end{algorithm}
\end{small}

\subsection{Batch Process Implementation}

Suppose that the training images are annotated into $K$ categories and each category contains a number $O$ of images. We can thus obtain a maximum number $K*O*(O-1)*(K-1)* O$ of triplet samples, which is cubically more than the source images. Since the number of stored images possibly reaches to millions in practice, it is hence expected to avoid loading all the data at once. To this end, we implement the model training in a batch-process fashion. Specifically, in each round, only a small set of triplets is produced and fed to the neural networks. However, randomly producing triplets is infeasible, as it may lead to the fact that the image distribution over the triplets is scattered and any two triplets have very small possibility sharing the same image. This fact will make the valid training samples very few and further degenerate the pairwise comparison optimization. To overcome this issue, we present an efficient yet effective triplet generation scheme, which involves the following steps in each iteration. We first randomly choose $\widehat{K}$ semantic categories, from which a number $\widehat{O}$ of images are   randomly selected. Then, for each selected image $I_k$, we construct a fixed number of triplets, and in each triplet the image having different label from $I_k$ is randomly selected from the remaining categories. In this way, the images distributed over the generated triplet samples are relatively centralized, so that we can collect more pairwise label information for learning. Moreover, since the categories and images are selected randomly for each iteration, this generation method will produce all possible triplet samples with a large enough number of iterations. In all of our experiments, we set $\widehat{K} = 10$ and $\widehat{O} = 20$.

\section{Experiments}
\label{sec:Experiment}

\subsection{Dataset and Experimental Setting}

We validate our deep hashing learning framework on several public datasets of image retrieval, including \textbf{MNIST}\footnote{http://yann.lecun.com/exdb/mnist/}, \textbf{CIFAR-10}\footnote{http://www.cs.toronto.edu/~kriz/cifar.html}, \textbf{CIFAR-20}\footnote{http://www.cs.toronto.edu/~kriz/cifar.html} and \textbf{NUS-WIDE}\footnote{http://lms.comp.nus.edu.sg/research/NUS-WIDE.htm}. For each dataset, the images are split into a training set and a query set. We use the training set to learn the network parameters and use the query set to compare the competing methods. Note that, in all of the experiments, the query image is searched within the query set itself by applying the leave-one-out procedure. Moreover, we evaluate our hashing method in the application of person re-identification using \textbf{CHUK03}\cite{Person-ReIdentification} dataset.

Several variants of our framework are evaluated in experiments. For notation simplicity, we denote our framework as DRSCH (\textit{i.e.}, Deep Regularized Similarity Comparison Hashing). To justify our formulation, we implement one simplified variant of our framework, namely DSCH, by removing the Laplacian regularization term. Note that both DRSCH and DSCH do not have the element-wise layer illustrated in Fig. \ref{fig:framework} and output the binary hash code with specified length directly. To analyze the effectiveness of different components of the end-to-end framework, we further remove the tanh-like layer to evaluate their influence to the final results. The output of this model is continuous and the algorithm returns the ranking list according to the Euclidean distance. Without special instruction, we will use "Euclidean" to indicate this model. Table~\ref{table:tab1}$\sim$\ref{table:tabadd4} show the results of the ranking measure in different dataset. The bit-scalable versions of DRSCH and DSCH are denoted by BS-DRSCH and BS-DSCH, respectively and the evaluation of these two methods will be reported in Sec.~\ref{sec:Bit-Scalable}. We compare our methods with eight state-of-the-art approaches:

\begin{itemize}
\item[1)] Locality Sensitive Hashing (LSH)~\cite{DBLP:LSH}: LSH generates a set of random linear projection as hash functions. We adopt the Gaussian random matrix as the set of hash functions, each column of which indicates a special random projection. The same setting is used  in~\cite{DBLP:ITQ}\cite{DBLP:Sparse-EmbeddingHashing-TIP}.
\item[2)] Spectral Hashing (SH)~\cite{DBLP:SH}: SH first employs PCA on the original data, then calculate the analytical Laplacian eigenfunctions along the principal directions. Hash codes are generated according to the projection based on these eigenfunctions.
\item[3)] Iterative Quantization (ITQ)~\cite{DBLP:ITQ}: ITQ is also a PCA-based hashing method which first conducts PCA on the original data and then finds an orthogonal matrix to make the variance of each bit maximized and hash bits pairwise uncorrelated.
\item[4)] PCA-Random Rotation (PCA-RR)~\cite{DBLP:ITQ}: PCA-RR is the basic version of ITQ, which adopts the random orthogonal matrix instead of learning based orthogonal matrix proposed in ITQ.
\item[5)] Minimal Loss Hashing (MLH)~\cite{DBLP:MLH}: By treating the hash code as the latent variables, MLH adopts the structured prediction formulation for hash learning. Based on binary hashing loss-adjusted inference and perceptron-like learning, an online efficient learning algorithm is employed for the optimization of hash functions.
\item[6)] Binary Reconstructive Embedding (BRE)~\cite{DBLP:BRE}: BRE does not require any assumptions on data distribution, and directly learns the hash functions by minimizing the reconstruction error between the distances in the original feature space and the Hamming distances in the embedded binary space.
\item[7)] Kernel-based Supervised Hashing (KSH)~\cite{DBLP:KSH}: KSH is a kernel based method which maps the data to binary hash codes by maximizing the separability of code inner products between similar and dissimilar pairs. Different from DRSCH, KSH adopts the kernel trick to learn nonlinear hash functions on the hand-crafted feature space.
\item[8)] Deep Semantic Ranking Hashing (DSRH)~\cite{DBLP:DeepSemanticRankingHash}:  DSRH is a recent developed method that incorporates feature learning into hash learning framework to preserve multilevel semantic similarity between multi-label images.
\end{itemize}

The first four methods are unsupervised and the others are supervised methods. The experimental results of first seven methods are obtained by the released implementations provided by their authors with the suggested feature representations and parameters provided in their papers. For fair comparison, we further evaluate three hashing methods (\textit{i.e.,} KSH-CNN, MLH-CNN and BRE-CNN) on the features extracted from the activation of last fully-connected layer of the neural network (\textit{i.e.}, AlexNet~\cite{ImageNetClassification}) pre-trained on the ImageNet\footnote{http://www.image-net.org/} dataset. In this way, CNN can be seen as a generic feature generator~\cite{DBLP:DeepSemanticRankingHash}\cite{DBLP:NouralCode}. The last compared approach is DSRH which is also based on the deep learning framework. Since the source code of DSRH~\cite{DBLP:DeepSemanticRankingHash} is not released, we carefully implement DSRH and our approach based on Caffe\footnote{http://caffe.berkeleyvision.org/} and obtain the final results. Note that the network parameters of DSRH~\cite{DBLP:DeepSemanticRankingHash} and our method are initialized randomly without any pre-training.

To evaluate the hashing methods, we utilize two search procedures, \textit{i.e.}, Hamming ranking and hash lookup~\cite{DBLP:Semi-SupervisedHashing-PAMI}\cite{DBLP:Sparse-EmbeddingHashing-TIP}. Hamming ranking gives the ranking list for all images in the database based on their Hamming distance or Hamming affinity to the query, where the ideal semantic neighbors are expected to be returned on the top of the ranking list. Hash lookup constructs a lookup table, and all the points in the buckets that fall into a small Hamming radius of the query are returned \cite{DBLP:Semi-SupervisedHashing-PAMI}. In our experiments, three Hamming ranking and one Hash lookup performance metrics are adopted. (1) Mean Average Precision (MAP) \cite{MAP}. Since the calculation of MAP is inefficient for large dataset, following \cite{DBLP:Sparse-EmbeddingHashing-TIP}, we report the results of top 50K returned neighbors for NUS-WIDE. (2) \textit{precision@500}, \textit{i.e.}, the average precision of the first 500 returned image for each query with different lengths of hash codes. (3) \textit{precision@k}, \textit{i.e.}, the fraction of \textit{k} closest images that are from the same-class or with semantic consistency in a certain Hamming space. (4) HAM2, \textit{i.e.}, the precision curve with the Hamming distance between the query image and dataset smaller than 2. The first three metrics evaluate the performance of Hamming ranking and the last one evaluates the result of Hash lookup. These four metrics reflect the different properties of hashing methods. The higher the values of all these four metrics are, the better the performance is.


\subsection{Network and Parameter Setting}

In the proposed framework, we resize the images to size $64\times64$ for the NUS-WIDE dataset, and resize the input images of MNIST, CIFAR10 and CIFAR20 to $28\times28$, $32\times32$ and $32\times32$ respectively. The parameter $\lambda$ in Eq.(\ref{eq_Lapmodel}) is set as $0.001$ in all the experiments. In each iteration, we load 10 semantic categories images (for NUS-WISE the batch is selected according to the semantic tags but not class labels), each of which includes about 20 images. So in total 200 images are feed into the network in each iteration, and they will generate about 684,000 triplets for training. In order to accelerate the training process, we randomly select 200,000 triplets to calculate the gradient. Note that the similarity matrix $S$ in Eq.~(\ref{eq_Lapmodel}) is also constructed according to the selected images in each iteration, and thus our method avoids constructing the overall similarity matrix and it is scalable to large scale dataset.

\begin{table}
\renewcommand{\arraystretch}{1.1}
\addtolength{\tabcolsep}{-1pt}
\begin{center}
\begin{tabular}{c|cccccc}
\hline
\multicolumn{1}{c|}{\multirow {2}{*}{Method}} & \multicolumn{5}{c}{MNIST (MAP \%)} \\
\cline{2-6}
        & 16 bits & 24 bits & 32 bits & 48 bits & 64 bits  \\
\hline
\textbf{DRSCH}   & \textbf{96.92} & \textbf{97.37}  & \textbf{97.88}  & \textbf{97.91}  & \textbf{98.09}              \\
\textbf{DSCH }                                       & 96.51& 96.63  & 97.21  & 97.48  & 97.68                \\
DSRH~\cite{DBLP:DeepSemanticRankingHash}             & 96.48 & 96.69  &  97.21 &  97.53 &   97.75            \\
KSH-CNN~\cite{DBLP:KSH}                              & 83.89 & 86.67  & 88.51  & 89.41  &  89.67             \\
MLH-CNN~\cite{DBLP:MLH}                              & 71.03 & 76.18  & 78.06  & 80.66  &  80.87             \\
BRE-CNN~\cite{DBLP:BRE}                              & 61.00 & 64.05  & 64.11  & 66.33  &  67.02             \\
KSH~\cite{DBLP:KSH}                                  & 82.85 & 86.03  & 87.37  & 88.48  & 88.82              \\
MLH~\cite{DBLP:MLH}                                  & 45.77 & 62.16  & 63.07  & 65.23  & 66.70                \\
BRE~\cite{DBLP:BRE}                                  & 41.96 & 57.19  & 56.52  & 64.74  & 66.55              \\
PCA-RR~\cite{DBLP:ITQ}                               & 35.96 & 39.93  & 38.17  & 43.81  & 45.76              \\
ITQ~\cite{DBLP:ITQ}                                  & 34.44 & 38.99  & 40.62  & 43.04  & 41.76            \\
SH~\cite{DBLP:SH}                                    & 13.40 & 14.81  & 15.28  & 16.29  & 17.11               \\
LSH~\cite{DBLP:LSH}                                  & 22.65 & 21.39  & 35.56  & 27.85  & 37.78              \\
\hline
Euclidean                                            & 89.55 & 87.83  & 86.89  & 83.76  & 82.92            \\
\hline
\end{tabular}
\vspace{1em}
\caption{Image retrieval results (Mean Average Precision) with various number of bits on the MNIST dataset. The scale of test query set is 10K. Our method outperforms the state-of-the-art methods.}\label{table:tab1}
\end{center}
\end{table}

\begin{table}
\renewcommand{\arraystretch}{1.1}
\addtolength{\tabcolsep}{-1pt}
\begin{center}
\begin{tabular}{c|cccccc}
\hline
\multicolumn{1}{c|}{\multirow {2}{*}{Method}} & \multicolumn{5}{c}{CIFAR-10 (MAP \%)}\\
\cline{2-6}
                & 16 bits & 24 bits & 32 bits & 48 bits & 64 bits \\
\hline
\textbf{DRSCH }                  & \textbf{61.46} & \textbf{62.19}   & \textbf{62.87} & \textbf{63.05 }   &\textbf{63.26 }     \\
\textbf{DSCH}                     & 60.87  & 61.33 & 61.74 & 61.98  &  62.35     \\
DSRH~\cite{DBLP:DeepSemanticRankingHash}     & 60.84 & 61.08  & 61.74  & 61.77  &  62.91             \\
KSH-CNN~\cite{DBLP:KSH}  & 40.08 & 42.98  & 44.39  & 45.77  &  46.56             \\
MLH-CNN~\cite{DBLP:MLH}  & 25.04 & 28.86  & 31.29  & 31.88  &  31.83             \\
BRE-CNN~\cite{DBLP:BRE}  & 19.80 & 20.57  & 20.59  & 21.64  &  21.96             \\
KSH~\cite{DBLP:KSH}                     & 32.15  & 35.17   & 36.51   & 38.26   & 39.50  \\
MLH~\cite{DBLP:MLH}                     & 13.33  & 15.78   & 16.29   & 18.03   & 18.84   \\
BRE~\cite{DBLP:BRE}                     & 12.19  & 15.63   & 16.10   & 17.19   & 17.56  \\
PCA-RR~\cite{DBLP:ITQ}                  & 12.06  & 12.24   & 13.61   & 13.46   & 13.80  \\
ITQ~\cite{DBLP:ITQ}                     & 11.45  & 11.63   & 11.53   & 10.97   & 11.24  \\
SH~\cite{DBLP:SH}                      & 19.22  & 19.28   & 20.09   & 20.79   & 21.46  \\
LSH~\cite{DBLP:LSH}                     & 12.36  & 11.74   & 12.30   & 13.57   & 12.42  \\
\hline
Euclidean                               & 35.46 & 34.07  & 33.91  &  32.18 & 31.09      \\
\hline
\end{tabular}
\vspace{1em}
\caption{Image retrieval results (Mean Average Precision) with various number of bits on the CIFAR-10 dataset. The scale of test query set is 10K (1K per class). The proposed method outperforms the state-of-the-art methods. }\label{table:tab2}
\end{center}
\end{table}

\begin{table}
\renewcommand{\arraystretch}{1.1}
\addtolength{\tabcolsep}{-1pt}
\begin{center}
\begin{tabular}{c|cccccc}
\hline
\multicolumn{1}{c|}{\multirow {2}{*}{Method}} & \multicolumn{5}{c}{NUS-WIDE (MAP \%)}\\
\cline{2-6}
                & 16 bits & 24 bits & 32 bits & 48 bits & 64 bits \\
\hline
\textbf{DRSCH}                      & \textbf{61.81}  & \textbf{62.24}  & 62.27  &62.79  & \textbf{64.14} \\
\textbf{DSCH }                    & 59.17  & 59.74  & 61.05  & 60.89  & 62.76 \\
DSRH~\cite{DBLP:DeepSemanticRankingHash}     & 60.92 & 61.78  & 62.13  & \textbf{63.09 } &  64.02             \\
KSH-CNN~\cite{DBLP:KSH}  & 60.74 & 61.89  & \textbf{62.46}  & 62.57  &  63.11             \\
MLH-CNN~\cite{DBLP:MLH}  & 52.51 & 55.91  & 56.83  & 58.07  &  59.79             \\
BRE-CNN~\cite{DBLP:BRE}  & 53.80 & 55.79  & 56.58  & 57.58  &  59.13             \\
KSH~\cite{DBLP:KSH}                     & 54.56  & 55.63  & 56.22  & 56.68  & 58.35   \\
MLH~\cite{DBLP:MLH}                     & 48.71  & 50.69  & 51.11  & 52.38  & 54.03   \\
BRE~\cite{DBLP:BRE}                     & 48.64  & 51.45  & 51.83  & 52.75  & 54.66   \\
PCA-RR~\cite{DBLP:ITQ}                 & 42.15  & 40.39  & 41.94  & 42.68  & 44.57   \\
ITQ~\cite{DBLP:ITQ}                     & 45.23  & 46.14  & 46.71  & 47.07  & 47.29  \\
SH~\cite{DBLP:SH}                      & 43.33  & 43.26  & 43.81  & 43.06  & 45.18  \\
LSH~\cite{DBLP:LSH}                     & 40.18  & 41.88  & 42.26  & 43.04  & 45.48  \\
\hline
Euclidean                              & 48.85 &  48.23 &  47.93 & 47.06  &  46.79   \\
\hline
\end{tabular}
\vspace{1em}
\caption{Image retrieval results (Mean Average Precision) with various number of bits on the NUS-WIDE dataset. The scale of test query set is 2100 (100 images for each semantic label). Our method achieves the competing performance compared with the state-of-the-art methods . }\label{table:tab3}
\end{center}
\end{table}

\begin{table}
\renewcommand{\arraystretch}{1.1}
\addtolength{\tabcolsep}{-1pt}
\begin{center}
\begin{tabular}{c|cccccc}
\hline
\multicolumn{1}{c|}{\multirow {2}{*}{Method}} & \multicolumn{5}{c}{CIFAR-20 (MAP \%)}\\
\cline{2-6}
                & 16 bits & 24 bits & 32 bits & 48 bits & 64 bits \\
\hline
\textbf{DRSCH }                  & \textbf{23.41} & \textbf{23.79}   & \textbf{24.38} & \textbf{25.63}   &\textbf{26.51}     \\
\textbf{DSCH}                           & 22.64 & 23.07 & 23.88  &  24.16   &  24.67         \\
DSRH~\cite{DBLP:DeepSemanticRankingHash}& 22.71 & 23.39  & 23.86  & 24.05  & 24.74              \\
KSH-CNN~\cite{DBLP:KSH}                 & 18.53 & 19.89  & 21.23  & 23.11  & 23.87              \\
MLH-CNN~\cite{DBLP:MLH}                 & 10.94 & 12.09  & 12.89  & 14.36  & 15.33              \\
BRE-CNN~\cite{DBLP:BRE}                 & 9.98 &  10.67 & 11.16  & 11.44  & 11.95              \\
KSH~\cite{DBLP:KSH}                     & 9.11 &  9.42  & 9.99   & 10.36  & 10.92\\
MLH~\cite{DBLP:MLH}                     & 7.15  & 7.32  & 7.45   & 7.85   & 8.10  \\
BRE~\cite{DBLP:BRE}                     & 7.33  & 7.62  & 7.62   & 8.01   & 8.11  \\
\hline
Euclidean                               &13.92  & 11.86  & 11.41  &  10.95 & 10.97      \\
\hline
\end{tabular}
\vspace{1em}
\caption{Image retrieval results (Mean Average Precision) with various number of bits on the CIFAR-20 dataset. The scale of test query set is 10K (500 per class). Our DRSCH outperform the state-of-the-art methods with obvious margins. }\label{table:tabadd4}
\end{center}
\end{table}

\begin{figure*}[htbp] \centering
\subfigure[] { \label{fig:mnist_hamm_dist}
\begin{minipage}[c]{0.28\linewidth}
\centering
\includegraphics[height=1.8in]{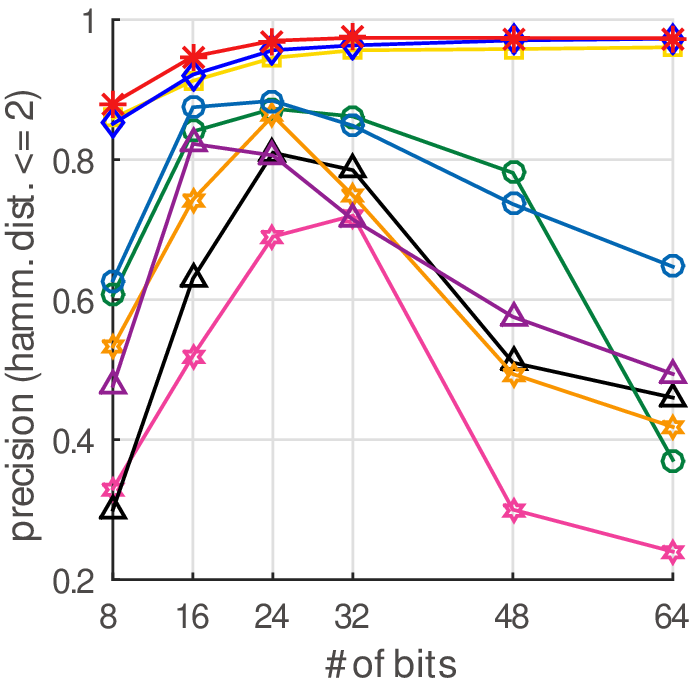}
\end{minipage}%
}%
\hspace{-0.0in}
\subfigure[] { \label{fig:mnist_precision_500}
\begin{minipage}[c]{0.28\linewidth}
\centering
\includegraphics[height=1.8in]{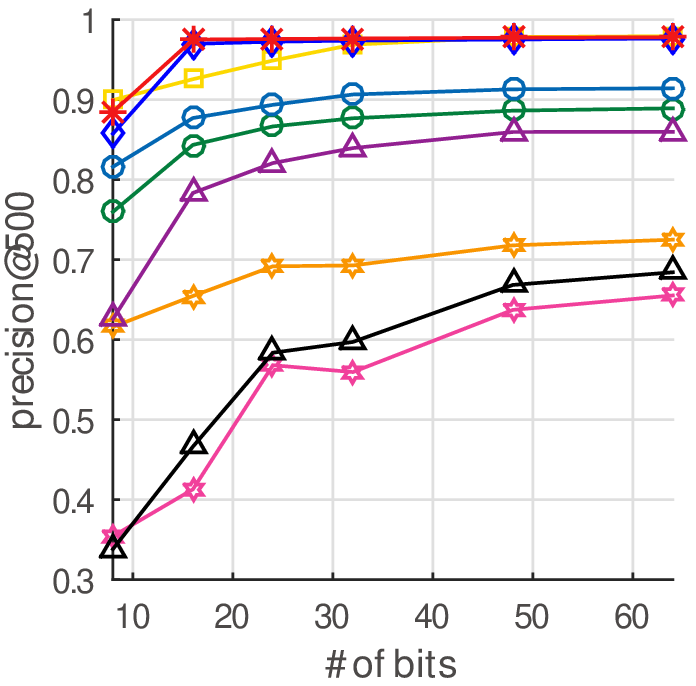}
\end{minipage}%
}%
\hspace{-0.0in}
\subfigure[] { \label{fig:mnist_precision_k}
\begin{minipage}[c]{0.35\linewidth}
\centering
\includegraphics[height=1.8in]{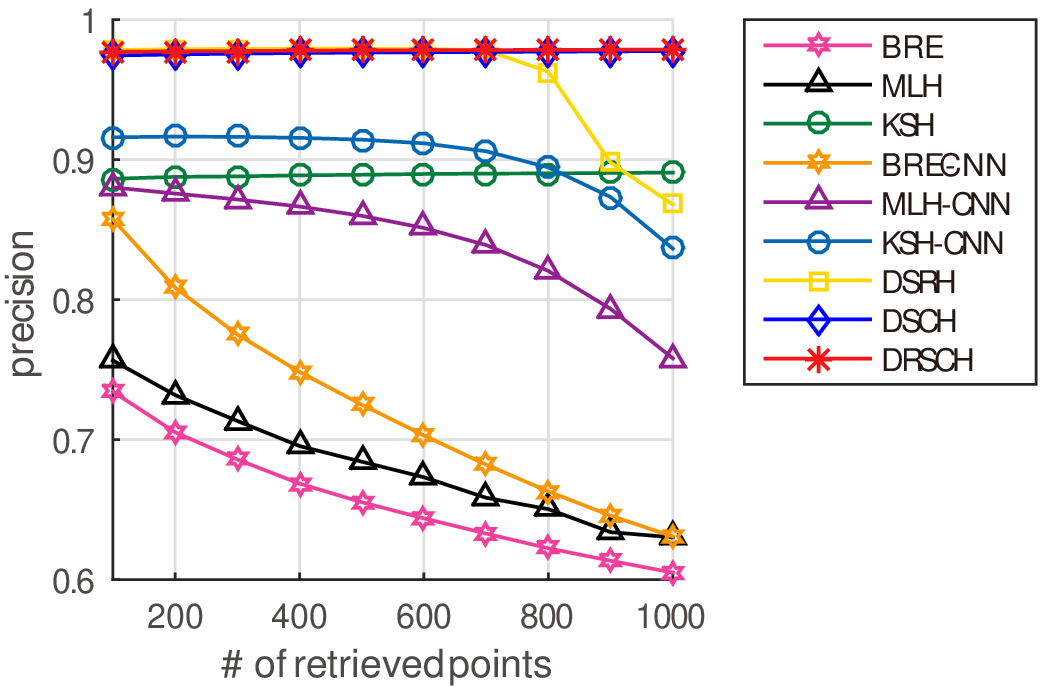}
\end{minipage}
}
\caption{ The results on the MNIST dataset. (a) Precision curves within Hamming radius 2; (b) Precision curves with top 500 returned; (c) Precision curves with 64 hash bits.}
\label{fig:mnist}
\end{figure*}

\begin{figure*}[htbp] \centering
\subfigure[] { \label{fig:cifar10_hamm_dist}
\begin{minipage}[c]{0.28\linewidth}
\centering
\includegraphics[height=1.8in]{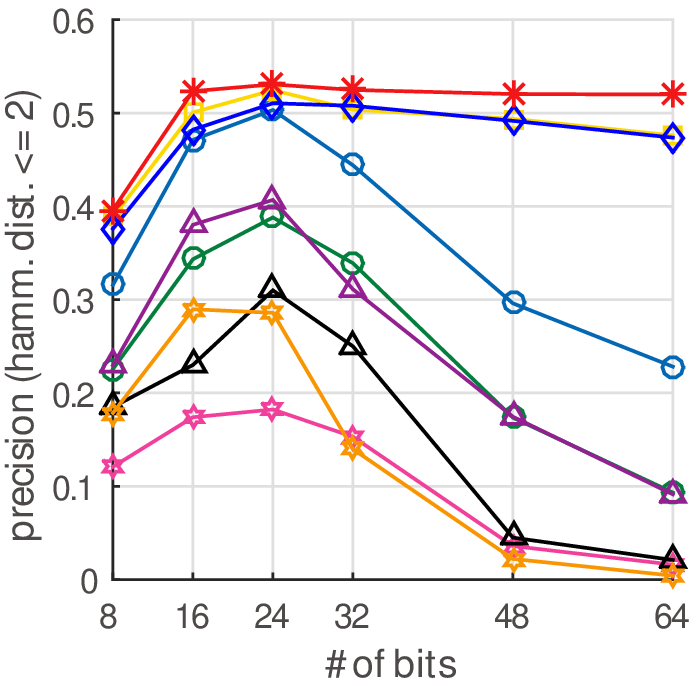}
\end{minipage}%
}%
\hspace{-0.0in}
\subfigure[] { \label{fig:cifar10_precision_500}
\begin{minipage}[c]{0.28\linewidth}
\centering
\includegraphics[height=1.8in]{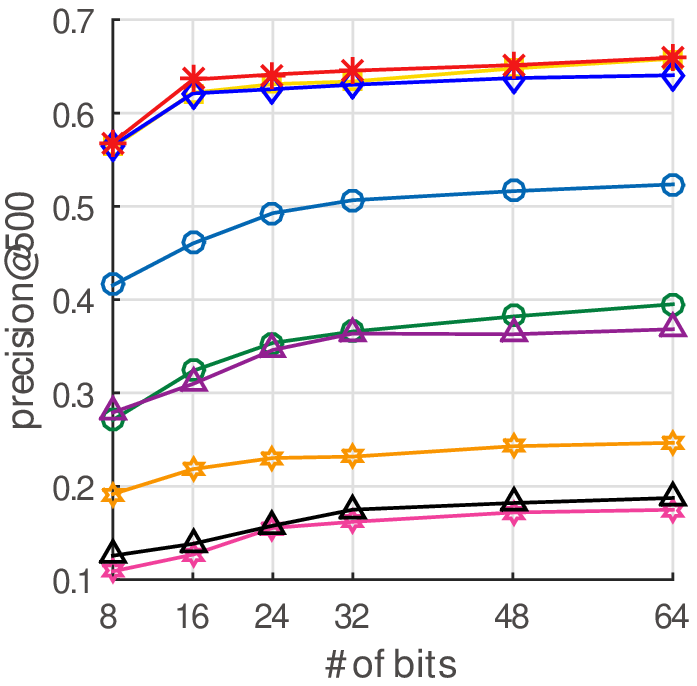}
\end{minipage}%
}%
\hspace{-0.0in}
\subfigure[] {  \label{fig:cifar10_precision_k}
\begin{minipage}[c]{0.35\linewidth}
\centering
\includegraphics[height=1.8in]{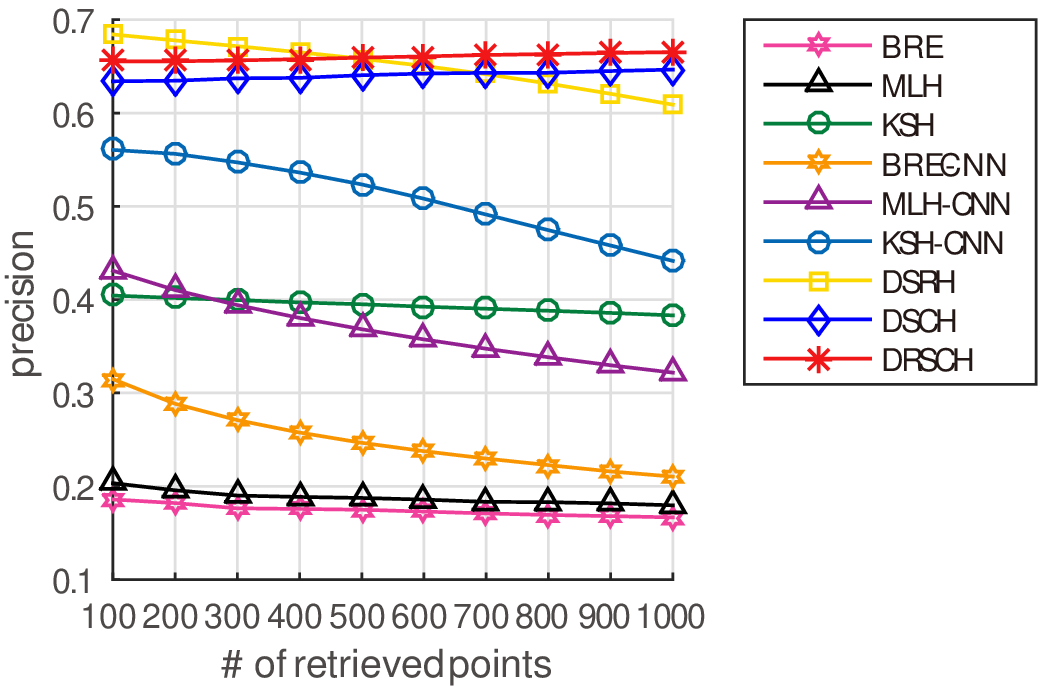}
\end{minipage}
}
\caption{ The results on the CIFAR-10 dataset. (a) Precision curves within Hamming radius 2; (b) Precision curves with top 500 returned; (c) Precision curves with 64 hash bits. }
\label{fig:CIFAR}
\end{figure*}

\begin{figure*}[htbp] \centering
\subfigure[] { \label{fig:nus_hamm_dist}
\begin{minipage}[c]{0.28\linewidth}
\centering
\includegraphics[height=1.8in]{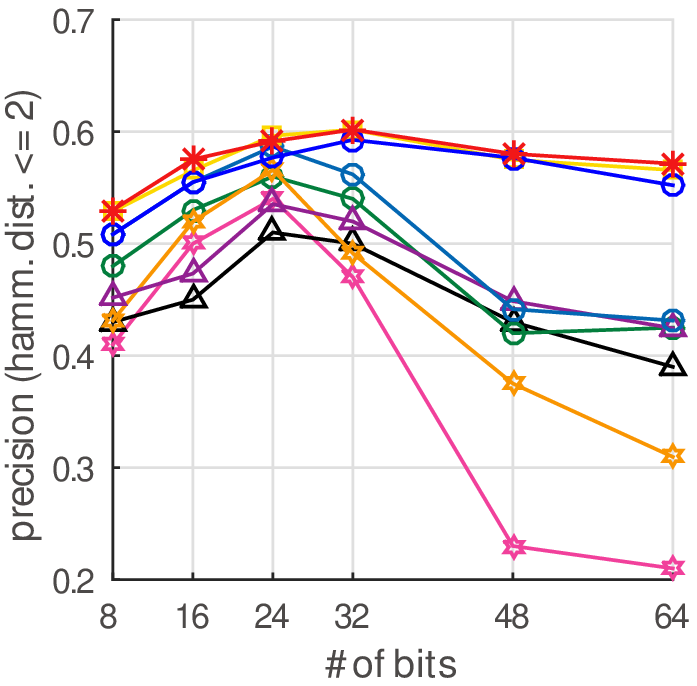}
\end{minipage}%
}%
\hspace{-0.0in}
\subfigure[] { \label{fig:nus_precision_500}
\begin{minipage}[c]{0.28\linewidth}
\centering
\includegraphics[height=1.8in]{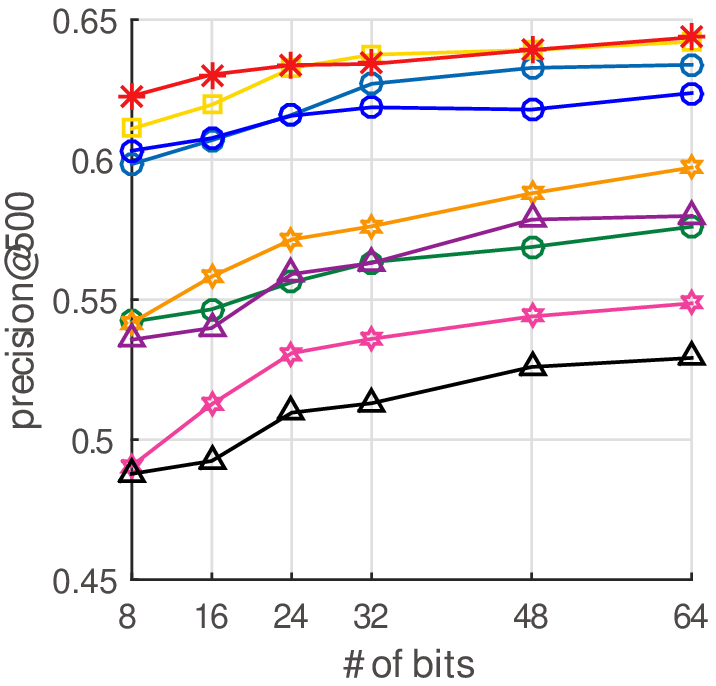}
\end{minipage}%
}%
\hspace{-0.0in}
\subfigure[] { \label{fig:nus_precision_k}
\begin{minipage}[c]{0.35\linewidth}
\centering
\includegraphics[height=1.8in]{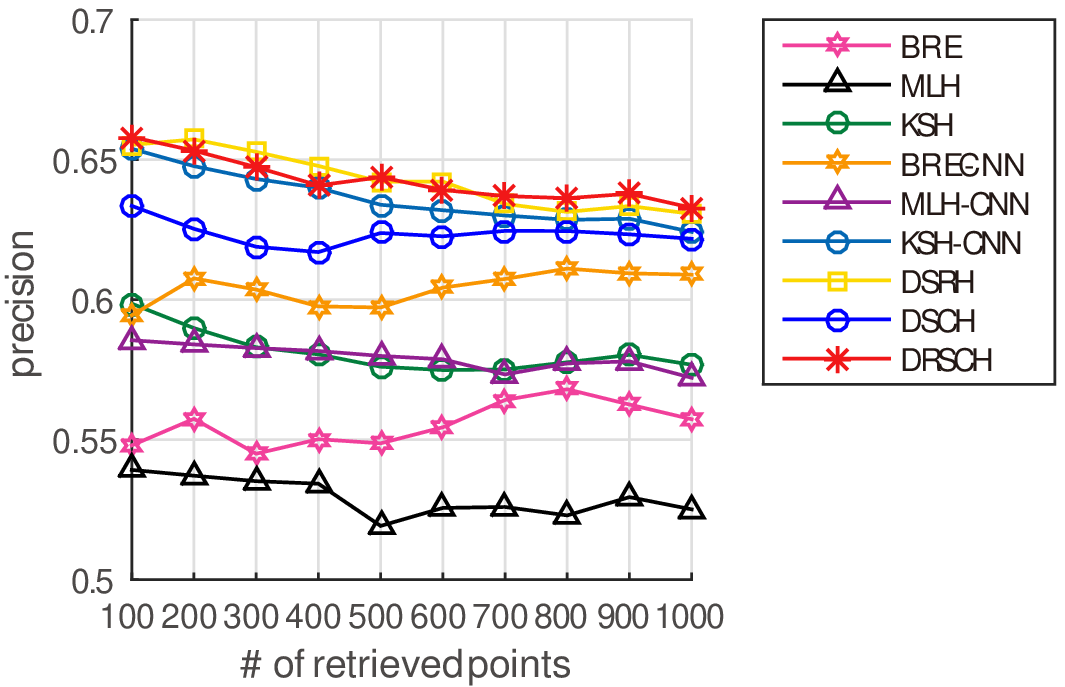}
\end{minipage}
}
\caption{ The results on the NUS-WIDE dataset. (a) Precision curves within Hamming radius 2; (b) Precision curves with top 500 returned; (c) Precision curves with 64 hash bits.}
\label{fig:NUS}
\end{figure*}

\begin{figure*}[htbp] \centering
\subfigure[] { \label{fig:cifar20_hamm_dist}
\begin{minipage}[c]{0.28\linewidth}
\centering
\includegraphics[height=1.8in]{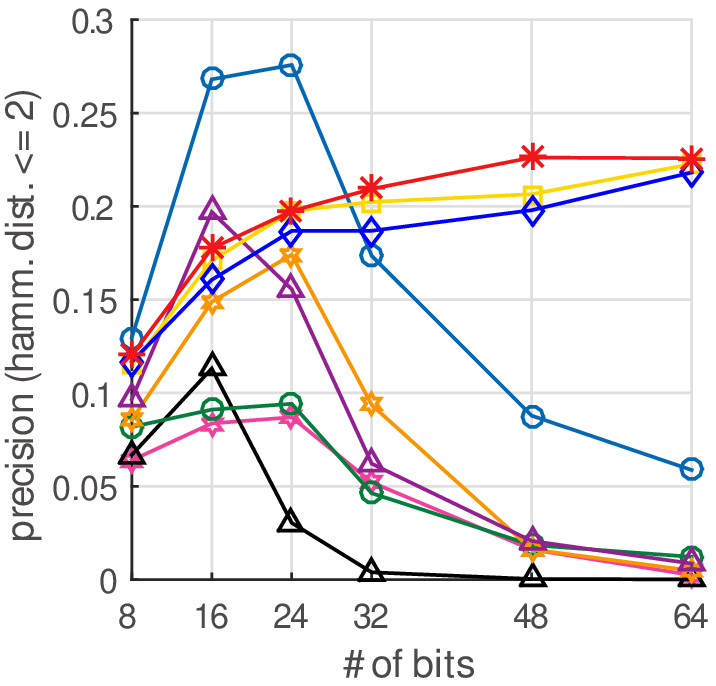}
\end{minipage}%
}%
\hspace{-0.0in}
\subfigure[] { \label{fig:cifar20_precision_500}
\begin{minipage}[c]{0.28\linewidth}
\centering
\includegraphics[height=1.8in]{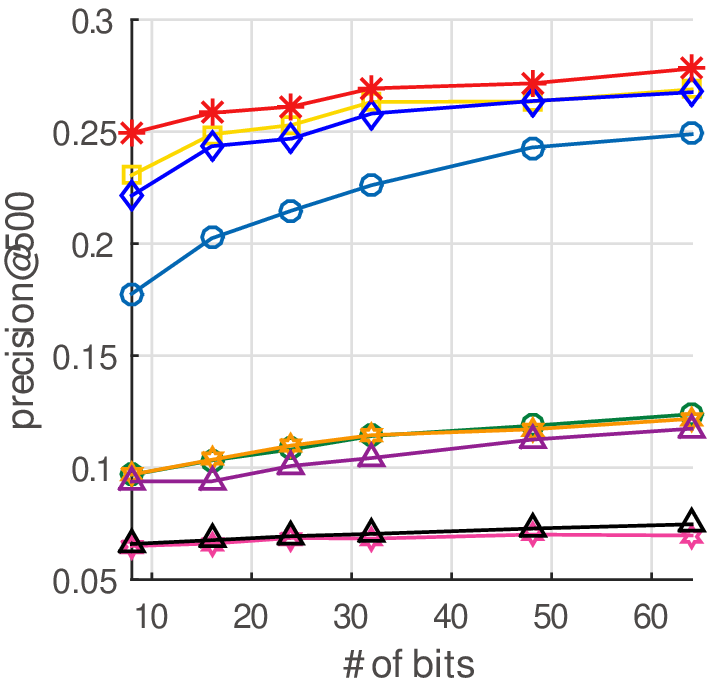}
\end{minipage}%
}%
\hspace{-0.0in}
\subfigure[] {  \label{fig:cifar20_precision_k}
\begin{minipage}[c]{0.35\linewidth}
\centering
\includegraphics[height=1.8in]{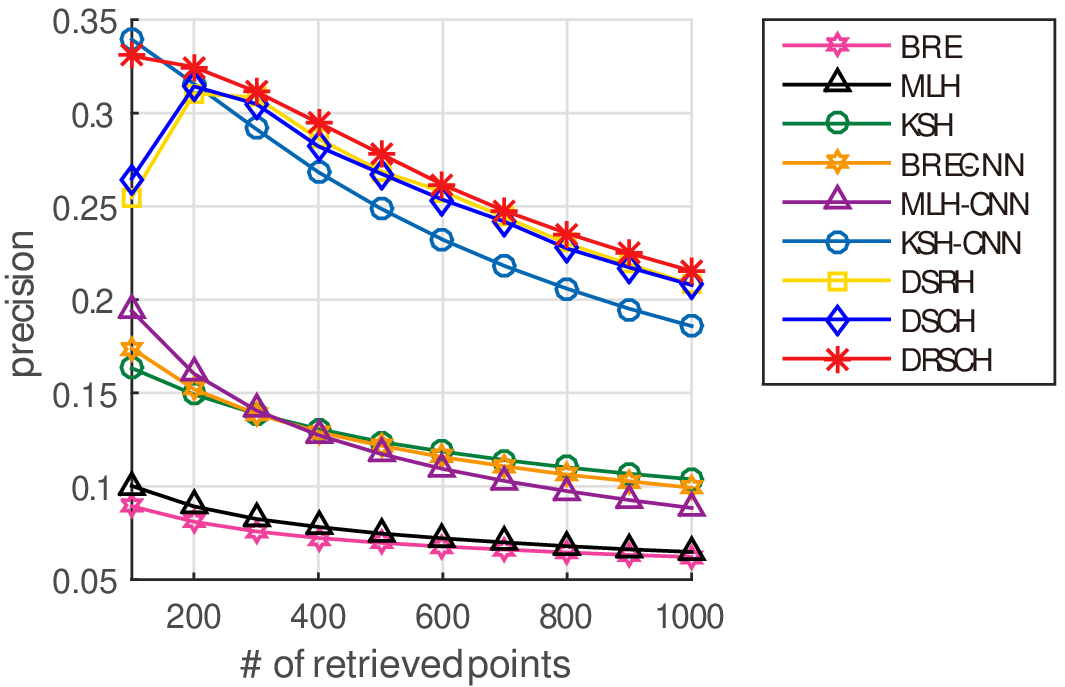}
\end{minipage}
}
\caption{ The results on the CIFAR-20 dataset. (a) Precision curves within Hamming radius 2; (b) Precision curves with top 500 returned; (c) Precision curves with 64 hash bits. }
\label{fig:CIFAR20}
\end{figure*}

\subsection{Experiments on Benchmark Dataset}

\begin{table*}
\renewcommand{\arraystretch}{1.1}
\addtolength{\tabcolsep}{-1pt}
\begin{center}
\begin{tabular}{c|c|c|c|c|c|c|c|c|c}
\hline
\multicolumn{1}{c|}{\multirow {2}{*}{Method}} & \multicolumn{1}{c|}{\multirow {2}{*}{Processing Unite}} & \multicolumn{2}{c|}{MNIST (ms)} & \multicolumn{2}{c|}{CIFAR-10 (ms)} & \multicolumn{2}{c|}{NUS-WIDE (ms)} & \multicolumn{2}{c}{CIFAR-20 (ms)}\\
\cline{3-10}
          &              & H + S & F + H + S & H + S & F + H + S  & H + S & F + H + S & H + S & F + H + S\\
\hline
\textbf{DRSCH}                                 & CPU \& GPU     & --   & 2.223   & --    & 3.257   &   --     &  3.566    &  --  &  3.408    \\
DSRH~\cite{DBLP:DeepSemanticRankingHash}       & CPU \& GPU     & --   & 4.745   & --    & 6.510   &  --      &  6.492    &  --  &  6.586    \\
KSH-CNN~\cite{DBLP:KSH}                        & CPU \& GPU     & 2.098& 6.499   & 2.172 & 6.754   &   2.168  &  6.613    & 2.112&  6.744    \\
KSH-Fea.~\cite{DBLP:KSH}                       & CPU            & 0.428& 0.664   & 0.556 & 175.782 &   0.501  &  177.863  & 0.488&  175.694  \\
MLH-CNN~\cite{DBLP:MLH}                        & CPU \& GPU     & 1.269& 5.669   & 1.298 & 5.898   &   1.273  &  5.718    & 1.242&  5.842    \\
MLH-Fea.~\cite{DBLP:MLH}                       & CPU            & 1.081& 1.317   & 1.202 & 176.428 &   1.267  &  178.629  & 1.227&  176.473   \\
BRE-CNN~\cite{DBLP:BRE}                        & CPU \& GPU     & 2.156& 6.656   & 2.229 & 6.809   &   2.414  &  6.859    & 2.341&  6.972    \\
BRE-Fea.~\cite{DBLP:BRE}                       & CPU            & 0.379& 0.615   & 0.547 & 175.773 &   0.513  &  177.875  & 0.487&  175.693  \\
\hline
\end{tabular}
\vspace{1em}
\caption{Comparison of the average testing time (millisecond per image) on four benchmark datasets by fixing the code length 64. For each traditional method, the suffix -fea. and -CNN denote the hand-craft feature and CNN feature respectively. }\label{table:tabtime}
\end{center}
\end{table*}

\textbf{Experiment I: MNIST}

We first report the performance of DSCH and DRSCH on handwritten digit retrieval by MNIST, which is one of the most popular datasets to test hashing methods \cite{DBLP:MLH,DBLP:LLH}. MNIST contains 70K greyscale handwritten digital images from "0" to "9" and each image has $28\times28$ pixels. Following the experiment setting in \cite{DBLP:LLH}, we use 10K images as the query set and the other 60K as the training samples. The pairwise similarity matrix $S$ in Eq. (\ref{eq_regTerm}) is constructed according to the class labels (\textit{i.e.}, the value corresponding to the image pair from the same class is set to one and zero otherwise.) For the method in~\cite{DBLP:DeepSemanticRankingHash} and our proposed DSCH and DRSCH, we directly apply the raw pixels as the input. For the other competing methods, we apply 784 dimensional vector (\textit{i.e.}, $28\times28$) as the traditional feature representation~\cite{DBLP:MLH}. And 4096 dimensional vector is extracted from AlexNet~\cite{ImageNetClassification} as the deep feature representation.

Fig.~\ref{fig:mnist_hamm_dist} shows the precision curve within Hamming distance 2 for different lengths of hash bits (\textit{i.e.}, from 8-bits to 64-bits). Fig.~\ref{fig:mnist_precision_500} reports the Precision$@500$ for different code lengths. Fig.~\ref{fig:mnist_precision_k} illustrates the Precision$@k$ utilizing 64-bit binary codes on MNIST. The MAP results with different code lengths are listed in Table~\ref{table:tab1}. Our DRSCH and DSCH outperform all of the other methods in all cases. In particular, DRSCH has at least $10\%$ gain over traditional methods even with CNN features under all code lengths, which demonstrates the benefit of joint optimization rather than the classical cascaded scheme (\textit{i.e.},  feature extraction followed by hashing). The performance of raw CNN feature (without tanh-like layer), which is also provided in Table~\ref{table:tab1}, indicates our hash functions are coherent with the deep feature representation.



\textbf{Experiment II: CIFAR-10}

The CIFAR-10 dataset consists of 60K $32\times32$ color images from 10 classes, with 6K images per class. We randomly sample 10K query images (1K images per object class) and use the rest as the training set. The similarity matrix $S$ is constructed based on the category labels as well. For fair comparison, each image is represented by the 512-dimensional GIST feature vector~\cite{DBLP:KSH} and 4096-dimensional CNN feature representation respectively.

Fig.~\ref{fig:cifar10_hamm_dist} shows image retrieval results within Hamming distance 2 for different hash bits; Fig.~\ref{fig:cifar10_precision_500} shows the Precision$@500$ results; and  Fig.~\ref{fig:cifar10_precision_k} illustrates the Precision$@k$ obtained using 64-bit binary codes. Table~\ref{table:tab2} gives the MAP results with different code lengths. Although the CNN features boost the performance of traditional cascade methods by a obvious margin, our approach still outperforms these methods because of joint optimization of the feature representation and hash functions. It also achieves relative increase of $1.67\%$ compared with DSRH (the deep learning method)~\cite{DBLP:DeepSemanticRankingHash} (one state-of-the-art deep hashing method) .

\textbf{Experiment III: NUS-WIDE}

The NUS-WIDE dataset collects about 270K images associated with 81 semantic labels from the web. Different from MNIST and CIFAR-10 where each sample has a unique class label, NUS-WIDE is a multi-label dataset where each image is annotated with one or multiple concept labels.  Following \cite{DBLP:Sparse-EmbeddingHashing-TIP}, we only consider the 21 most frequently happened semantic labels and the number of associated images is $195,969$. We randomly sample 100 images from each of the 21 semantic categories as queries and use the rest as training samples. The matching groundtruth is defined as a pair of images that share at least one common label. We construct the similarity matrix $S$ based on the proportion of shared labels:
\begin{equation}\label{eq-21}
    S_{ij} = \frac{\mathcal{s}_{i}\bigcap \mathcal{s}_{j}}{\mathcal{s}_{i}\bigcup \mathcal{s}_{j}},
\end{equation}
where $S_{ij}$ denotes the semantic similarity of images $i$ and $j$, $\mathcal{s}_i$ and $\mathcal{s}_j$ denote the semantic label set of image $i$ and image $j$, respectively. We adopt 512-dimensional GIST vector and 4096-dimensional CNN vector as image feature representations for traditional approaches and resize each image into $64\times64$ for our DSCH and DRSCH.

The precision curve within Hamming distance 2, the Precision$@500$ for varied code lengths and the Precision$@k$ utilizing 64-bit binary codes are reported in Fig.~\ref{fig:nus_hamm_dist}, Fig.~\ref{fig:nus_precision_500} and Fig.~\ref{fig:nus_precision_k}, respectively. For NUS-WIDE, two images are regarded as semantically similar if they share at least one label. Table~\ref{table:tab3} lists the results of different hash learning methods under the MAP metric. Since NUS-WIDE is very large, we just calculate the MAP within the first 50K searched neighbors.


\textbf{Experiment IV: CIFAR-20}

Just like CIFAR-10, CIFAR-20 is another famous dataset for object recognition and image retrieval, which contains 20 superclasses grouped from CIFAR-100 dataset. For each class there are 2500 training images and 500 testing images. To compare with the traditional hashing learning method with hand-crafted feature, each image is represented by GIST vector with the feature dimension 512. Following~\cite{DBLP:NouralCode}, we also extract 4096-dimensional CNN feature as generic visual representation for further comparison.

Fig.~\ref{fig:cifar20_hamm_dist} shows image retrieval results within Hamming distance 2 for different hash bits; Fig.~\ref{fig:cifar20_precision_500} shows the Precision$@500$ results; and  Fig.~\ref{fig:cifar20_precision_k} illustrates the Precision$@k$ obtained using 64-bit binary codes. Table~\ref{table:tabadd4} gives the MAP results with different code lengths and our DRSCH still works the best. However, with scale of the dataset growing, the achieved performance gain becomes insignificant. One of the reasonable explanation is that the benefit of the joint optimization degrades at such scales. This is because the classes are much more populated and the manifold distribution is much more complicated to estimate by triplet based comparison in such scale.

\subsection{Efficiency Analysis}

All the experiments are carried out on a PC with NVIDIA Tesla K40 GPU, Intel Core i7-3960X 3.30GHZ CPU and 24GB memory. The average testing time of our approach and competing methods on four benchmark datasets are reported in Table~\ref{table:tabtime}. For simplicity, we use capital letter ``F'', ``H'' and ``S'' to indicate feature extraction, hash code generation and image search respectively. For all the experiments, we assume every image in the database has already been represented by the binary hash code. In this way, the time consumption of feature extraction and hash code generation are mainly caused by the query image. Since the forward propagation of the neural network only needs a series of matrix multiplication and convolution operations and can be efficiently computed with GPU (Graphics Processing Unit) implementation, it is obvious that our DRSCH is relatively slow when the competing methods ignore the time cost of feature extraction. In contrast, when feature extraction is taking into consideration, efficiency will be a distinct advantage of our end-to-end framework. Actually, for traditional cascaded methods, calculating the generic feature costs 99\%(for 512-dimensional Gist feature) of testing time. In this case, our CNN-based hashing can be more efficient than those cascaded ones. Note that the cascade methods are performed on the raw pixels as features on MNIST dataset, making them slightly more efficient than our DRSCH.

\begin{table}
\renewcommand{\arraystretch}{1.1}
\addtolength{\tabcolsep}{-1pt}
\begin{center}
\begin{tabular}{c|ccccccc}
\hline
\multicolumn{1}{c|}{\multirow {2}{*}{Method}} & \multicolumn{6}{c}{MNIST (MAP \%)} \\
\cline{2-7}
              & 8 bits & 16 bits & 24 bits & 32 bits & 48 bits & 64 bits  \\
\hline
DRSCH         & 91.69& 96.92& 97.37  & 97.88  & 97.91  & 98.09              \\
DSCH          & 90.38& 96.51& 96.63  & 97.21  & 97.48  & 97.68               \\
BS-DRSCH     & 94.11 & 96.91& 97.15 & 97.36 & 97.39  & 97.35               \\
\hline
\end{tabular}
\vspace{1em}
\caption{Image retrieval results (Mean Average Precision) with various number of bits on the MNIST dataset. The size of the test query set is 10K.}\label{table:tab4}
\end{center}
\end{table}

\begin{table}
\renewcommand{\arraystretch}{1.1}
\addtolength{\tabcolsep}{-1pt}
\begin{center}
\begin{tabular}{c|ccccccc}
\hline
\multicolumn{1}{c|}{\multirow {2}{*}{Method}} & \multicolumn{6}{c}{CIFAR-10 (MAP \%)}\\
\cline{2-7}
               & 8 bits & 16 bits & 24 bits & 32 bits & 48 bits & 64 bits \\
\hline
DRSCH           & 58.92  & 62.46 & 62.19   & 62.87  & 63.05    & 63.26     \\
DSCH            & 57.17  & 60.87 & 61.33   & 61.74  & 61.98    & 62.35     \\
BS-DRSCH       & 58.03 & 61.37 & 62.29   & 62.53  & 62.75    & 62.81     \\
\hline
\end{tabular}
\vspace{1em}
\caption{Image retrieval results (Mean Average Precision) with various number of bits on the CIFAR-10 dataset. The size of the test query set is 10K (1K per class).}\label{table:tab5}
\end{center}
\end{table}

\begin{table}
\renewcommand{\arraystretch}{1.1}
\addtolength{\tabcolsep}{-1pt}
\begin{center}
\begin{tabular}{c|ccccccc}
\hline
\multicolumn{1}{c|}{\multirow {2}{*}{Method}} & \multicolumn{6}{c}{NUS-WIDE (MAP \%)}\\
\cline{2-7}
              & 8 bits  & 16 bits & 24 bits & 32 bits & 48 bits & 64 bits \\
\hline
DRSCH          & 55.71 & 61.81  & 62.24  & 62.27  & 62.79 & 64.14 \\
DSCH           & 53.25 & 59.17  & 59.74  & 61.05  & 60.89 & 62.76 \\
BS-DRSCH      & 58.77 & 62.05  & 62.41  & 62.64  & 63.33 & 63.82  &                \\
\hline
\end{tabular}
\vspace{1em}
\caption{Image retrieval results (Mean Average Precision) with various number of bits on the NUS-WIDE dataset. The size of the test query set is 2100.}\label{table:tab6}
\end{center}
\end{table}

\begin{table}
\renewcommand{\arraystretch}{1.1}
\addtolength{\tabcolsep}{-1pt}
\begin{center}
\begin{tabular}{c|ccccccc}
\hline
\multicolumn{1}{c|}{\multirow {2}{*}{Method}} & \multicolumn{6}{c}{CIFAR-20 (MAP \%)}\\
\cline{2-7}
              & 8 bits  & 16 bits & 24 bits & 32 bits & 48 bits & 64 bits \\
\hline
DRSCH          & 22.31 & 23.41  &  23.79 &  24.38 & 25.63 & 26.51 \\
DSCH           & 20.01 & 22.64 &  23.07 &  23.88 & 24.16 & 24.67 \\
BS-DRSCH      &  22.98 &  24.63 & 24.81  & 24.84  & 24.85 & 25.14  \\
\hline
\end{tabular}
\vspace{1em}
\caption{Image retrieval results (Mean Average Precision) with various number of bits on the CIFAR-20 dataset. The size of the test query set is 10K (0.5K per class).}\label{table:tab7}
\end{center}
\end{table}

\subsection{Evaluation of Bit-Scalable Hashing}
\label{sec:Bit-Scalable}

In this subsection, we evaluate the performance of the proposed Bit-Scalable Deep Hashing method. In the training phase, BS-DRSCH is used to learn a weighted hash code with the maximum bit length ({\em i.e.}, $q$ = 64). In the test phase, for any length of hash code $k$ $(k\leq q)$, we select the $k$ bits with the largest weights to calculate the Hamming similarity according to Eq.(\ref{eq_HamAff}). Therefore, BS-DRSCH is bit-scalable to hashing applications with any bit length.

\begin{figure}[htbp] \centering
\subfigure[] {
\begin{minipage}[c]{0.7\linewidth}
\centering
\includegraphics[width=1\textwidth]{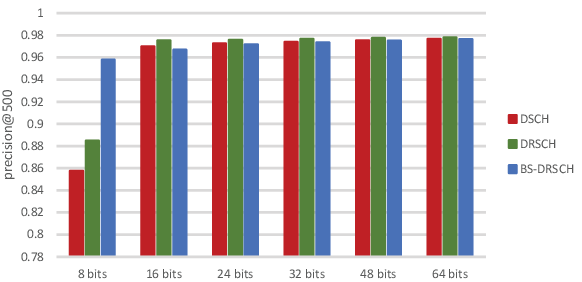}
\end{minipage}%
}%
\hspace{-0.0in}
\subfigure[] {
\begin{minipage}[c]{0.7\linewidth}
\centering
\includegraphics[width=1\textwidth]{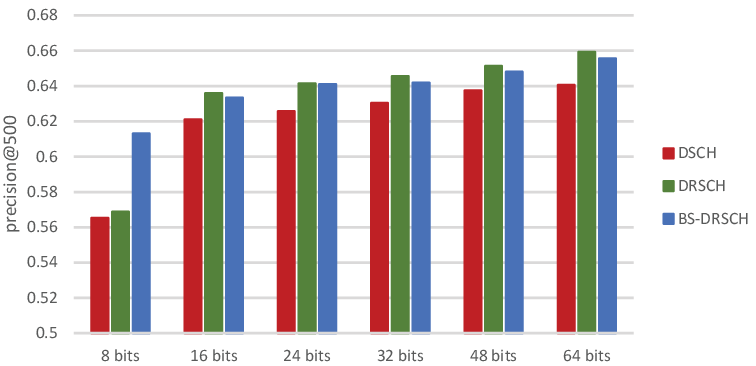}
\end{minipage}%
}%
\hspace{-0.0in}
\subfigure[] {
\begin{minipage}[c]{0.7\linewidth}
\centering
\includegraphics[width=1\textwidth]{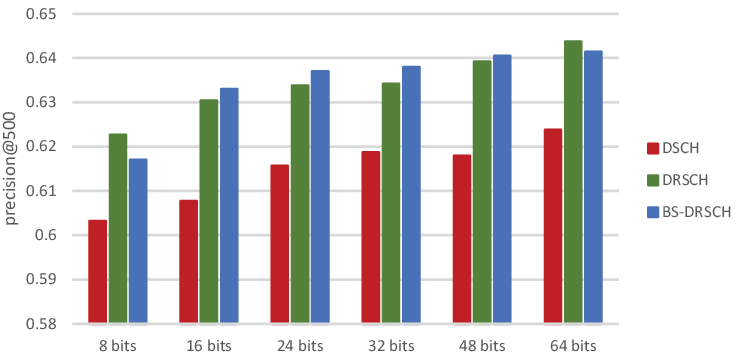}
\end{minipage}
}%
\hspace{-0.0in}
\subfigure[] {
\begin{minipage}[c]{0.7\linewidth}
\centering
\includegraphics[width=1\textwidth]{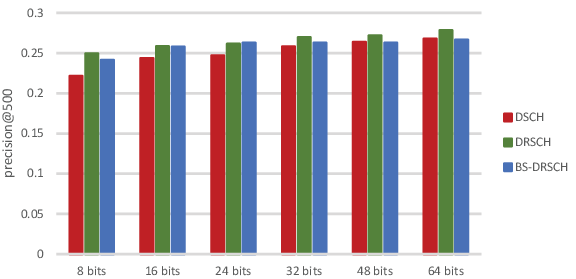}
\end{minipage}
}
\caption{ Precision@500 vs. \#bits. (a) MNIST dataset; (b) CIFAR-10 dataset; (c) NUS-WIDE dataset; (d) CIFAR-20 dataset}
\label{fig:comp}
\end{figure}

\begin{figure*}
\begin{center}
\includegraphics[width=1\textwidth]{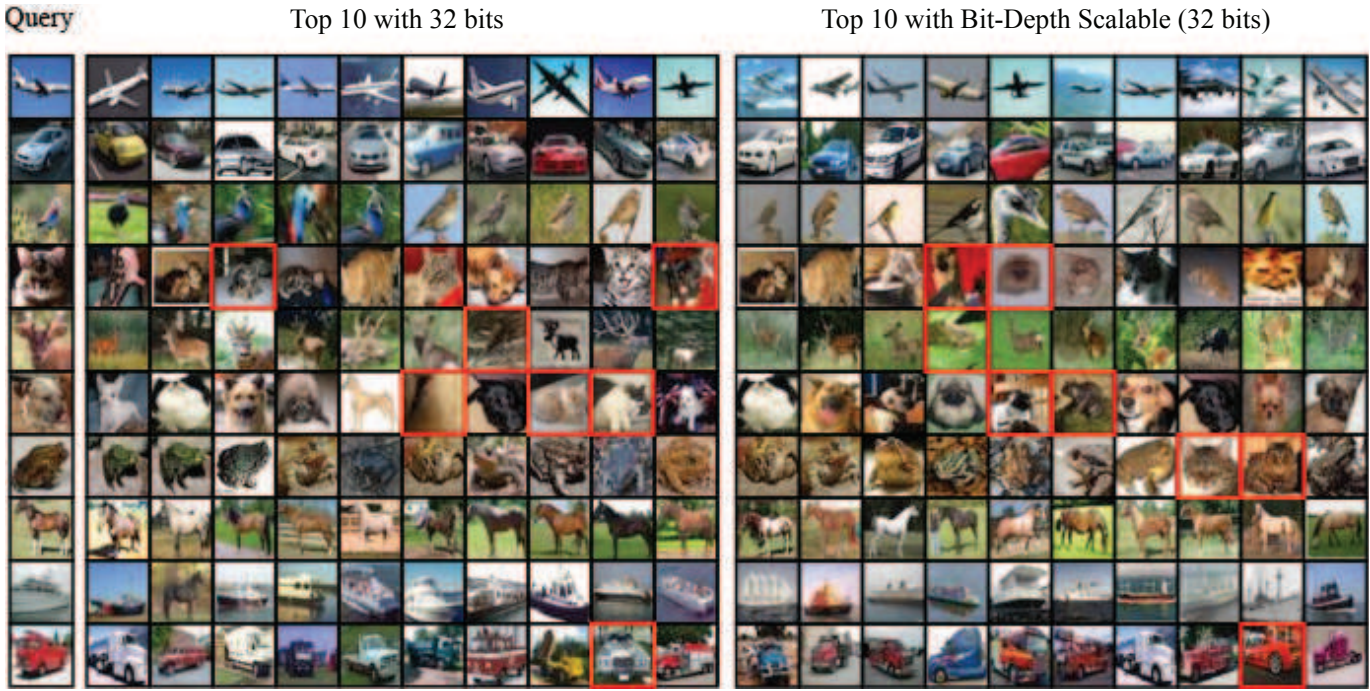}
\end{center}
   \caption{Retrieval results (top 10 returned images) for ten CIFAR-10 test images using Hamming ranking on 32-bit hash codes. The left column shows the query images. The middle 10 columns show the top returned images by fix length hashing learning algorithm. The right 10 columns indicate the top returned images adopting bit-scalable learning method. Red rectangles indicate mistakes. Note that for Bit-Scalable Hashing, we train a neural network with 64-bit output and select the 32 bits with the largest weights for testing.}
\label{fig:short}
\end{figure*}

The retrieval performance associated with various lengths of hash code is reported in Tables~\ref{table:tab4}$\sim$\ref{table:tab7}. It is obvious that BS-DRSCH achieves very competitive results with its fixed-length versions (\textit{i.e.}, DRSCH and DSCH ). The performances of precision@500 for different datasets are also reported in Fig.\ref{fig:comp} for further comparison. At last, Fig.\ref{fig:short} illustrates the retrieval results for ten CIFAR-10 test images by Hamming distance with 32-bit binary codes. From Tables V$\sim$VIII, when the number of bits is smaller (\textit{i.e.,}$\leq32$), BS-DRSCH generally outperforms DRSCH on MNIST, NUS-WIDE, and CIFAR-20. When the number of bits is larger, the performance gains would be insignificant. This might be explained by that weighted hash code could be approximated by non-weighted hash code with longer bits, and thus when the number of bits is sufficiently large, weighted and non-weighted hash codes would obtain similar performance. Note that BS-DRSCH only needs to train once, making BS-DRSCH very suitable to applications where varied lengths of hashing codes are required for different scenarios.

\subsection{Application to Person Re-Identification}

Person re-identification~\cite{Person-ReIdentification} at a distance across disjoint camera views is an important problem in intelligent video surveillance, particularly for the applications restricting the use of face recognition. It is also a foundation of threat detection, event understanding and many other surveillance applications. Despite considerable efforts been made, it is still an open problem due to the dramatic variations caused by different camera viewpoints and person pose changes. Here we apply our deep hashing for person re-identification as a preliminary attempt, and we will focus on this task in future work.

We evaluate our method using  \textbf{CUHK03}~\cite{Person-ReIdentification} dataset, which is one of current largest dataset for this task. It includes 13164 images of 1360 pedestrians collected from 6 different surveillance cameras. Each identity is observed by two disjoint camera views and has an average of 4.8 images in each view. Following~\cite{Person-ReIdentification}, the dataset is partitioned into training set (1160 persons), validation set (100 persons) and test set (100 persons). All the images are resized to $250\times100$. The pairwise similarity matrix in Eq.(\ref{eq_regTerm}) is constructed according to the person identity. The experiments are conducted with 10 random splits. We adopt the widely used Cumulative Matching Characteristic (CMC) curve~\cite{Person-ReIdentification} for quantitative evaluation and all the CMC curves indicate single-shot results.

We compare with three person re-identification methods (KISSME~\cite{KISSME}, eSDC~\cite{eSDC}, and FPNN~\cite{Person-ReIdentification}), four state-of-the-art hashing learning methods (BRE~\cite{DBLP:BRE}, MLH~\cite{DBLP:MLH}, KSH~\cite{DBLP:KSH} and DRSH~\cite{DBLP:DeepSemanticRankingHash}) and the Euclidean distance. For KISSME~\cite{KISSME} and eSDC~\cite{eSDC}, the experimental results are generated by their suggested feature representation and parameters setting. FPNN~\cite{Person-ReIdentification} is a deep learning based method and the validation set is adopted in this method to select parameters of the network. When using traditional hashing learning methods and Euclidean distance, the 4096 dimensional CNN features are extracted from pre-trained AlexNet as the input features. For DRSH~\cite{DBLP:DeepSemanticRankingHash} and our approach, parameters of the networks are learned from raw images without any pre-training.

Table \ref{table:tab222} reports the quantitative results generated by all of the competing methods. The hashing-based methods (including ours) perform using both 64 and 128 bits hashing codes, and the ranking list is based on the Hamming distance. Compared with state-of-the-arts of person re-identification, our deep hashing framework achieves the comparable performances and outperforms other hashing  methods with large margins on Rank-1 and Rank-5 identification rate.

\begin{table}
\renewcommand{\arraystretch}{1.1}
\addtolength{\tabcolsep}{-1pt}
\begin{center}
\begin{tabular}{c|cccccc}
\hline
\multicolumn{1}{c|}{\multirow {2}{*}{Method}} & \multicolumn{5}{c}{CUHK ( CMC \% )}\\
\cline{2-6}
              & TOP1  & TOP5 & TOP10  & TOP20 & TOP30 \\
\hline
\textbf{DRSCH-128 }                                   & 18.74 &  48.39 &  \textbf{69.66}   & \textbf{81.03} &  \textbf{91.28 }              \\
\textbf{DRSCH-64}                                     & \textbf{21.96} &  46.66 &  66.04 & 78.93   &  88.76               \\
DSRH-128~\cite{DBLP:DeepSemanticRankingHash}           & 8.05  &  26.10 &  45.82   & 64.95 &  79.03                 \\
DSRH-64~\cite{DBLP:DeepSemanticRankingHash}           & 14.44 &  43.38 &  66.77   & 79.19 &  87.45                  \\
KSH-CNN-128~\cite{DBLP:KSH}      &  3.65  &  11.71 &  19.75  & 30.68 &  43.46                 \\
KSH-CNN-64~\cite{DBLP:KSH}       &  3.12 &   12.90 &  19.96  & 32.59 &  45.62                 \\
MLH-CNN-128~\cite{DBLP:MLH}       & 2.75 &   11.62 &  24.61  & 39.68 &  49.26                 \\
MLH-CNN-64~\cite{DBLP:MLH}       &  1.75 &   8.14  &  19.6   & 35.64 &  47.45               \\
BRE-CNN-128~\cite{DBLP:BRE}       & 3.91 &   7.24  &  11.83  & 24.20 &  36.15                 \\
BRE-CNN-64~\cite{DBLP:BRE}       & 3.22 &  6.74    &  10.25  & 24.69 &  37.75                 \\
FPNN~\cite{Person-ReIdentification}             & 20.65 & \textbf{50.09} & 66.42  & 80.02 & 87.71 \\
KISSME~\cite{KISSME}           & 14.17 & 41.12 & 54.89  & 70.09 & 80.02 \\
eSDC~\cite{eSDC}             &  8.76 & 27.03 & 38.32  & 55.06 & 67.75 \\
Euclidean        &  6.03 & 19.83 & 29.93  & 45.22 & 57.35 \\
\hline
\end{tabular}
\vspace{1em}
\caption{Experimental results on CUHK03 dataset using manually labeled pedestrain bounding boxes. The evaluation is based on CMC approach}\label{table:tab222}
\end{center}
\end{table}

\section{Conclusion}
\label{sec:conclusion}

In this paper, we presented a novel bit-scalable hashing approach by integrating feature learning and hash function learning into a joint optimization framework via deep convolutional neural networks. A regularized similarity comparison formulation was introduced in the deep hashing learning framework to ensure image adjacency consistency, while an element-wise layer was designed to weigh the hashing codes so that bit-scalability can be easily obtained. Our approach demonstrated very promising results on standard image retrieval benchmarks, not only outperforming state-of-the-arts in terms of retrieval accuracy, but also greatly improving the flexibility of varied length hashing over existing approaches. There are several interesting directions along which we intend to extend this work. The first is to improve our framework by leveraging more semantics ({\em e.g.}, multiple attributes) of images. Another one is to introduce feedback learning in the framework, making it more powerful in practice.


%


%
%
%
%

\ifCLASSOPTIONcaptionsoff
  \newpage
\fi



%

\bibliographystyle{IEEEtran}
\bibliography{IEEEtran}

%

\begin{IEEEbiography}[{\includegraphics[width=1in,height=1.25in,clip,keepaspectratio]{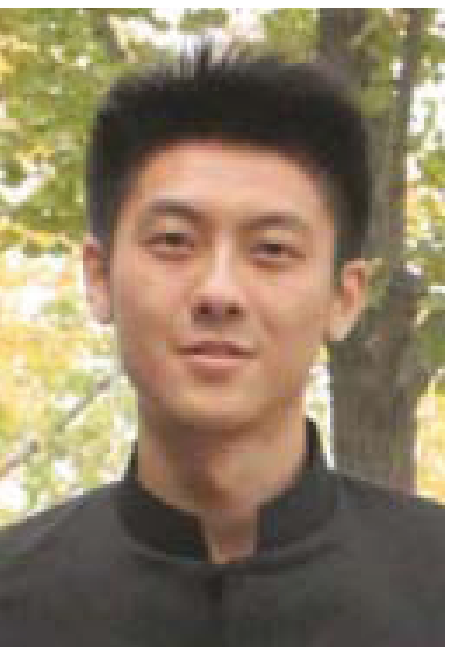}}]{Ruimao Zhang}
received the B.E. degree from the School of Software, Sun Yat-sen University, Guangzhou, China, in 2011, where he is currently working toward the Ph.D. degree in computer science with the School of Information Science and Technology. He was a Visiting Ph.D. Student with the Department of Computing, Hong Kong Polytechnic University, Hong Kong, from 2013 to 2014. His research interests include computer vision, pattern recognition, machine learning, and related applications.
\end{IEEEbiography}

\begin{IEEEbiography}[{\includegraphics[width=1in,height=1.25in,clip,keepaspectratio]{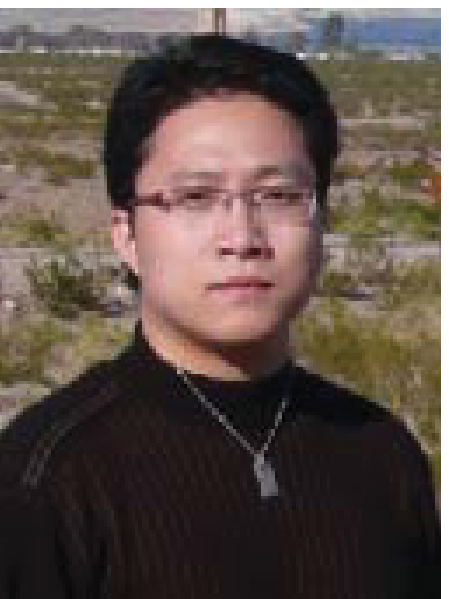}}]{Liang Lin}
is a Professor with the School of computer science, Sun Yat-Sen University (SYSU), China. He received the B.S. and Ph.D. degrees from the Beijing Institute of Technology (BIT), Beijing, China, in 1999 and 2008, respectively. From 2006 to 2007, he was a joint Ph.D. student with the Department of Statistics, University of California, Los Angeles (UCLA). He was a Post-Doctoral Research Fellow with the Center for Vision, Cognition, Learning, and Art of UCLA. His research focuses on new models, algorithms and systems for intelligent processing and understanding of visual data such as images and videos. He has published more than 80 papers in top tier academic journals and conferences. He currently serves as an associate editor of Neurocomputing and The Visual Computer. He was supported by several promotive programs or funds for his works such as Guangdong NSFs for Distinguished Young Scholars in 2013. He received the Best Paper Runners-Up Award in ACM NPAR 2010, Google Faculty Award in 2012, and Best Student Paper Award in IEEE ICME 2014.
\end{IEEEbiography}

\begin{IEEEbiography}[{\includegraphics[width=1in,height=1.25in,clip,keepaspectratio]{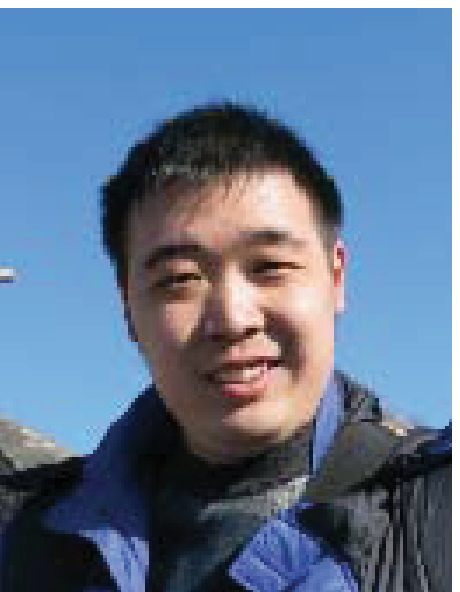}}]{Rui Zhang}
received B.E. degree from School of Software, Sun Yat-sen University, Guangzhou, China in 2014. He is currently working towards the MSc degree in software engineering with the School of Software. He was a research intern in IECA group, Microsoft Research Asia, from 2013 to 2014. His research interest includes computer vision, natural language processing, machine learning, and parallel computation.
\end{IEEEbiography}

\begin{IEEEbiography}[{\includegraphics[width=1in,height=1.25in,clip,keepaspectratio]{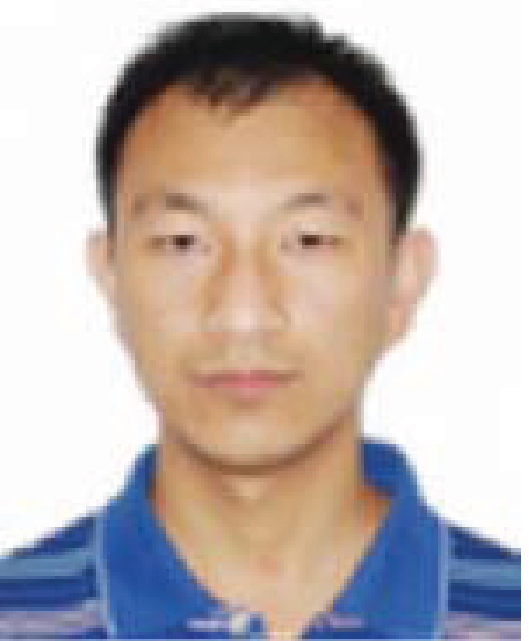}}]{Wangmeng Zuo}
(M'09, SM'14) received the Ph.D. degree in computer application technology from the Harbin Institute of Technology, Harbin, China, in 2007. In 2004, from 2005 to 2006, and from 2007 to 2008, he was a Research Assistant with the Department of Computing, Hong Kong Polytechnic University, Hong Kong. From 2009 to 2010, he was a Visiting Professor at Microsoft Research Asia. He is
currently an Associate Professor with the School of Computer Science and Technology, Harbin Institute of Technology. His current research interests include image modeling and low-level vision, discriminative learning, and biometrics. He has authored about 50 papers in those areas. He is an Associate Editor of the IET Biometrics.
\end{IEEEbiography}

\begin{IEEEbiography}[{\includegraphics[width=1in,height=1.25in,clip,keepaspectratio]{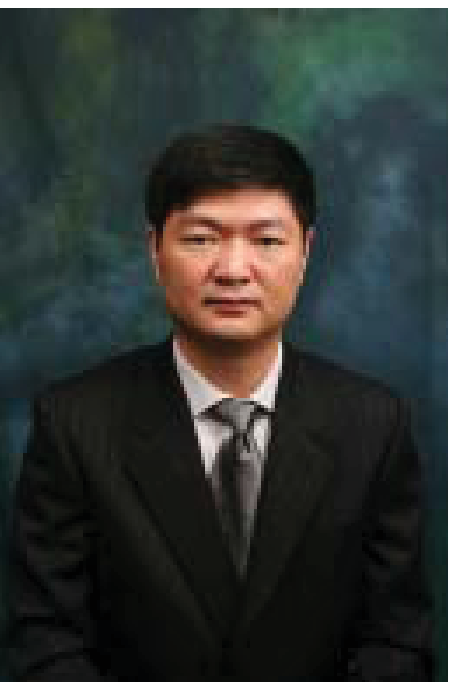}}]{Lei Zhang}
(M¡¯04, SM¡¯14) received the B.Sc. degree in 1995 from Shenyang Institute of Aeronautical Engineering, Shenyang, P.R. China, the M.Sc. and Ph.D degrees in Control Theory and Engineering from Northwestern Polytechnical University, Xi¡¯an, P.R. China, respectively in 1998 and 2001. From 2001 to 2002, he was a research associate in the Dept. of Computing, The Hong Kong Polytechnic University. From Jan. 2003 to Jan. 2006 he worked as a Postdoctoral Fellow in the Dept. of Electrical and Computer Engineering, McMaster University, Canada. In 2006, he joined the Dept. of Computing, The Hong Kong Polytechnic University, as an Assistant Professor. Since July 2015, he has been a Full Professor in the same department. His research interests include Computer Vision, Pattern Recognition, Image and Video Processing, and Biometrics, etc. Dr. Zhang has published more than 200 papers in those areas. By 2015, his publications have been cited more than 14,000 times in literature. Dr. Zhang is currently an Associate Editor of IEEE Trans. on Image Processing, IEEE Trans. on CSVT and Image and Vision Computing. He was awarded the 2012-13 Faculty Award in Research and Scholarly Activities. More information can be found in his homepage http://www4.comp.polyu.edu.hk/~cslzhang/.
\end{IEEEbiography}

\end{document}